\documentclass[sigconf]{acmart}
\usepackage{graphicx}
\usepackage{xcolor}
\usepackage{multirow}
\usepackage{tabularx}
\usepackage{algorithm}
\usepackage{algorithmic}
\usepackage{subcaption}
\usepackage{enumitem}
\usepackage{amsthm}

\copyrightyear{2021}
\acmYear{2021}
\setcopyright{acmcopyright}\acmConference[KDD '21]{Proceedings of the 27th ACM
SIGKDD Conference on Knowledge Discovery and Data Mining}{August 14--18,
2021}{Virtual Event, Singapore}
\acmBooktitle{Proceedings of the 27th ACM SIGKDD Conference on Knowledge
Discovery and Data Mining (KDD '21), August 14--18, 2021, Virtual Event, Singapore}
\acmPrice{15.00}
\acmDOI{10.1145/3447548.3467306}
\acmISBN{978-1-4503-8332-5/21/08}

\newcolumntype{C}{>{\centering\arraybackslash}X} 
\title{Labeled Data Generation with Inexact Supervision}

\author{Enyan Dai$^\dagger$, Kai Shu$^\ddagger$, Yiwei Sun$^\dagger$, Suhang Wang$^\dagger$ }

\affiliation{$\dagger$ The Pennsylvania State University,
{${\ddagger}$} Illinois Institute of Technology
}

\email{{emd5759, yus162, szw494}@psu.edu, kshu@iit.edu}

\begin{document}
\begin{abstract}
The recent advanced deep learning techniques have shown the promising results in various domains such as computer vision and natural language processing.
The success of deep neural networks in supervised learning heavily relies on a large amount of labeled data. However, obtaining labeled data with target labels is often challenging due to various reasons such as cost of labeling and privacy issues, which challenges existing deep models.
In spite of that, it is relatively easy to obtain data with \textit{inexact supervision}, i.e., having labels/tags related to the target task. For example, social media platforms are overwhelmed with billions of posts and images with self-customized tags, which are not the exact labels for target classification tasks but are usually related to the target labels. It is promising to leverage these tags (inexact supervision) and their relations with target classes to generate labeled data to facilitate the downstream classification tasks. However, the work on this is rather limited. Therefore, we study a novel problem of labeled data generation with inexact supervision. We propose a novel generative framework named as ADDES which can synthesize high-quality labeled data for target classification tasks by learning from data with inexact supervision and the relations between inexact supervision and target classes. 
Experimental results on image and text datasets demonstrate the effectiveness of the proposed ADDES for generating realistic labeled data from inexact supervision to facilitate the target classification task. 
\end{abstract}


\maketitle

\section{Introduction}
Deep learning technologies have achieved remarkable results in various domains such as image classification \cite{he2016deep}, object detection \cite{ren2015faster} and language translation \cite{gehring2017convolutional}.  However, training the deep neural networks relies on a large amount of labeled data, which is impractical to obtain in many domains. Taking the fake news detection for example, a news piece often requires hours of work from a professional to evaluate the credibility which leads to unaffordable time and labor costs. 
For applications in healthcare, it is difficult to obtain large-scale labeled data (e.g. EHR data) due to privacy issues and the scarcity of experts for labeling.

\begin{table}[t]
    \small
    \centering
    \caption{Samples of StackOverflow.}
    \label{tab:stackoverflow}
    \vskip -1.5em
    \begin{tabularx}{0.99\columnwidth}{p{0.46\columnwidth}p{0.20\columnwidth}X}
    \toprule
    Question Tittles & Tags & Target Labels \\
    \midrule
    Laravel 4: Input::all() returns no data with \$.ajax POST & ajax, laravel& php\\
    \midrule
    jquery DataTables.net plugin: how to ignore rows when sorting & jquery, html & javascript\\
    \bottomrule
    \end{tabularx}
    \vskip -2em
\end{table}

In spite of the difficulties in obtaining the accurately labeled data for target problems, the development of the internet and social media makes it easy to collect data with \textit{inexact supervision}, i.e., labels/tags related to the target classification task.
For example, users often post images and texts with self-customized tags on social media platforms such as Twitter, Facebook, and StackOverflow. Though these tags are often not the labels of target classes, they could provide inexact supervision through the relations between the tags and the labels of the target classification task.
Table \ref{tab:stackoverflow} gives two real examples of inexact supervision from StackOverflow, where questions are labeled with several tags and the target task is to assign the programming language to the questions based on their text. Obviously, these tags are not the target labels. However, they have relations with the labels of the target classes, which could provide inexact supervision for target label prediction. For instance, in the first example, the tag \texttt{laravel}  shows that the question is related with \texttt{php}, because \texttt{laravel} is a framework that designed for developing \texttt{php}. In the second example, the tag \texttt{jquery}, which is a JavaScript library, suggests that the text is likely to be related to the target label \texttt{javascript}. Thus, these tags could be used to help infer target labels even though no exact supervision (data with target labels) is given. 
There are various applications that could benefit from inexact supervision such as image classification for Flickr and short video classification for Instagram. Therefore, it is important to study learning from inexact supervision.

The recent development of deep generative models such as generative adversarial learning (GAN)~\cite{goodfellow2014generative} and variational autoencoder~\cite{kingma2013autoencoding} have shown promising results in generating realistic labeled data. The generated labeled data could be used to augment the dataset or facilitate the downstream classification tasks~\cite{antoniou2017data,shu2018deep,wang2020global,wang2020learning}. For example, 
data augmentation by GAN is shown effective for few-shot learning \cite{antoniou2017data}. \citeauthor{shu2018deep} \cite{shu2018deep} utilize the generated headlines for clickbait detection. Therefore, it is promising to develop deep generative models for generating labeled data from inexact supervision to facilitate the training of target classifiers. However, the work on labeled data generation with inexact supervision is rather limited.

Therefore, we investigate a novel problem of labeled data generation with inexact supervision. In essence, we aim to tackle the following challenges: (i) how to extract the information of the target labels from the labels of inexact supervision classes;  (ii) how to generate high-quality labeled data for classification. In an attempt to solve these two challenges, we propose a novel generative framework named as \textbf{ADDES}\footnote{https://github.com/EnyanDai/ADDES} (l\underline{a}bele\underline{d} \underline{d}ata generation with in\underline{e}xact \underline{s}upervision). To better infer the information for target label prediction, ADDES adopts a graph convolutional network (GCN) to capture the label relations. The information propagation among the nodes which represent different classes could utilize the supervision from labels of the inexact supervision classes. Furthermore, to obtain high-quality 
synthetic data, the framework is designed to utilize both the data with inexact supervision and unlabeled data. Our main contributions are:
\begin{itemize}[leftmargin=*]
    \item We study a novel problem of labeled data generation with inexact supervision for data augmentation;
    \item We propose a novel generative framework which could leverage data with inexact supervision, unlabeled data, and the relations between classes to generate  high-quality labeled data; and 
    \item  Experiments on the benchmark datasets, including the image and text datasets, demonstrate the effectiveness of our proposed framework for labeled data generation with inexact supervision.
\end{itemize}

The rest of the paper are organized as follows. In Section~\ref{sec:related_work}, we introduce related work. In Section~\ref{problem_definition}, we formally define the problem. In Section~\ref{sec:methodology}, we introduce the proposed method. In Section~\ref{sec:experiments}, we conduct experiments to demonstrate the effectiveness of the proposed method. In Section~\ref{sec:conclusion}, we conclude with future work.

\section{Related Work} \label{sec:related_work}



\subsection{Deep Generative Model}
Generative model aims to capture the distribution of the real data.
Recently, deep generative models such as generative adversarial networks (GANs)~\cite{goodfellow2014generative} and variational autoencoder (VAE)~\cite{kingma2013autoencoding}, have attracted increasing attention as a result of their strong power in generating realistic data samples. 
Based on GAN and VAE, various efforts~\cite{mirza2014conditional,odena2017conditional,sohn2015learning,bowman2015generating,hu2017toward,kingma2014semi} have been taken to generate realistic data with desired labels. For example, conditional GAN and conditional VAE are proposed to learn the conditional probability distribution of real data \cite{mirza2014conditional,sohn2015learning}. 
Controlled generation of text based on VAE is also explored \cite{bowman2015generating,hu2017toward}.
What's more, various applications of the generative models are investigated. One major application is to generate labeled data for data augmentation~\cite{wang2020global,wang2020learning,wang2018weakly,antoniou2017data,shu2018deep}. For example, data augmentation based on generative adversarial networks is explored in \cite{antoniou2017data}. In clickbait detection, headlines are generated to augment the data for better performance \cite{shu2018deep}.  In contrast to those prior works that require large-scale accurately labeled data to learn generative models for synthesizing labeled data, we investigate a new problem of generating labeled data without the ground truth of target labels. Moreover, the proposed framework ADDES is a unified framework that could effectively synthesize images and text.

\subsection{Learning from Weak Supervision}
For many real-world applications, obtaining large-scale high quality labels are difficult, while it is relatively easy weak supervision~\cite{zhou2018brief,ratner2017snorkel} such as noisy supervision~\cite{xiao2015learning,veit2017learning} and distant supervision~\cite{qin2018dsgan}. Thus, learning from weak supervision is attracting increasing attention and various approaches are proposed~\cite{xiao2015learning,veit2017learning,qin2018dsgan,han2019deep}. For example, \citeauthor{xiao2015learning}~\cite{xiao2015learning} model the relationships between images, class labels and label noises with a probabilistic graphical model and further integrate it into an end-to-end deep learning system. \citeauthor{han2019deep}~\cite{han2019deep} presents a novel deep
self-learning framework to train a robust network on the real noisy datasets without extra supervision. \citeauthor{qin2018dsgan}~\cite{qin2018dsgan} adopts generative adversarial training with distant supervision for relation extraction. Despite the various approaches for learning from weak supervision, the majority of them focus on noisy supervision and distant supervision.  The work on learning from inexact labels is rather limited, let labeled data generation from ineact labels.

\subsection{Multi-Label Classification}
Multi-label classification  is to predict a set of labels for an instance. The key challenge of multi-label learning is the overwhelming size of the possible label combinations. One straightforward way is to decompose the multi-label classification to a set of binary classification problems \cite{boutell2004learning}. To achieve better performance, researchers investigate a number of methods to capture the label dependencies. For example, \citet{wang2016cnn} use recurrent neural networks to model the high-order label dependency. \citet{chen_multi-label_2019} propose the ML-GCN to leverage the knowledge graph to explore the label correlation dependency. In our inexact supervision problem setting, we also assume that a data instance could have multiple labels. However, the multi-label learning assumes that all the labels of the instance are provided. We are dealing with a much more challenging problem that ground truth of target labels is totally missing in the training set, and our goal is to generate data of desired target labels for data augmentation.


\subsection{Zero-Shot Learning}

Zero-shot Learning (ZSL) aims to make classifications for the target categories when no data in these categories is provided. In this setting, only labeled data in the source categories is available for training. To transfer the knowledge learned from the source categories to the target categories, semantic embeddings such as word embeddings of the categories are utilized. Typical methods are to learn a compatibility function between the data and the semantic embeddings based on the source category images \cite{zhang2017learning,frome2013devise,romera2015embarrassingly}. Another direction is to sample features for the target categories from semantic embeddings through generative model \cite{Mishra_2018_CVPR_Workshops,Verma_2018_CVPR,xian2019f}. Recently, knowledge graph is adopted in zero-shot learning, which results in remarkable results \cite{lee_multi-label_2018,Wang_2018_CVPR,Kampffmeyer_2019_CVPR}. For example, \citet{Wang_2018_CVPR} build a graph linking the related categories together and use the GCN to predict the classifiers of the target categories from semantic embeddings. Although ZSL deals with the lack of labeled data in target categories, there is a distinct difference between zero-shot learning and inexact supervision learning. In zero-shot learning, an instance is supposed to belong to a single class. None of the supervision to the target category classification could be obtained from the seen categories data. On the contrary, we are interested in more practical scenarios in which seen labels of data could provide inexact supervision and be leveraged for target label prediction.

\section{Problem Definition} \label{problem_definition}
Let $\mathcal{S}=\{l_1,l_2,...,l_{|\mathcal{S}|}\}$ denotes the inexact supervision class set of size $|\mathcal{S}|$, and $\mathcal{T}=\{l_{|\mathcal{S}|+1},l_{|\mathcal{S}|+2},...,l_{|\mathcal{S}|+|\mathcal{T}|}\}$
denotes the target class set of size $|\mathcal{T}|$. Then the whole class set is  $ \mathcal{W} = \mathcal{S} \cup \mathcal{T}$. Note that the target class set has no overlap with the inexact supervision class set, i.e., $\mathcal{S} \cap \mathcal{T} = \emptyset$. 
The label vectors $\mathbf{y}_s \in \{0, 1\}^{|\mathcal{S}|}$ and $ \mathbf{y}_t \in \{0, 1\}^{|\mathcal{T}|}$ accordingly represent the labels of inexact supervision classes and target classes. For a data instance $x^k$, if $\mathbf{y}_s^k(i)==1$, it means the instance belongs to the class $l_i$, otherwise not. A data instance can belong to multiple classes. 
The whole training set $\mathcal{D}$ contains an inexact supervision class labeled data set $\mathcal{D}_l$ consisting of $n_l$ instances $(\mathbf{x}^l,\mathbf{y}_s^l)$ and an unlabeled data set $\mathcal{D}_u$ with $n_u$ unlabeled instances $\mathbf{x}^u$. The total training set can be written as:
\begin{equation} 
    \mathcal{D}=\mathcal{D}_l \cup \mathcal{D}_u= \{(\mathbf{x}^l, \mathbf{y}_s^l)\}^{n_l}_{l=1} \cup \{\mathbf{x}^u\}_{u=1}^{n_u},
\end{equation}
The labels from the whole class set $\mathcal{W}$ are correlated with each other. With the relations, the class graph $\mathcal{G} = \{\mathcal{W},\mathcal{E}\}$ is constructed, where $\mathcal{E} \subset \mathcal{W} \times \mathcal{W}$ is set of edges linking the classes.
We use $\mathbf{A} \in \mathbb{R}^{|\mathcal{W}| \times |\mathcal{W}|}$ to denote the correlation matrix. The weight $\mathbf{A}_{ij}$ indicates how likely the labels of classes $l_i$ and $l_j$ are both annotated to 1 in a single instance. The class embeddings matrix is denoted as $\mathbf{V} \in \mathbb{R}^{|\mathcal{W}| \times m}$, where $m$ is the dimension of the class embeddings. For example, the 
class embedding can be word embedding denoting the semantic meaning of the class or one hot encoding if word embedding is not available. With the notations and definitions described here, the problem of labeled data generation with inexact supervision for data augmentation could be formulated as:
\newtheorem{problem}{Problem}
\begin{problem}
Given the training set $\mathcal{D}=\mathcal{D}_l \cup \mathcal{D}_u$ and the graph $\mathcal{G}$ with the adjacency matrix $\mathbf{A}$ and class embeddings $\mathbf{V}$, we aim to learn a generative model and produce a set of labeled data $\mathcal{D}_s=\{(\mathbf{x}^{syn_i},\mathbf{y}_s^{syn_i},\mathbf{y}_t^{syn_i})\}^{n_s}_{i=1}$ through the following process:
\begin{equation}
    f(\mathbf{A},\mathbf{V}, \mathbf{y}_s^{syn},\mathbf{y}_t^{syn}) \rightarrow \mathbf{x}^{syn},
\end{equation}
where $f$ is the generative model required to learn. 
\end{problem}

\section{methodology} \label{sec:methodology}

\begin{figure*}[t]
    \centering
    \includegraphics[width=0.92\linewidth]{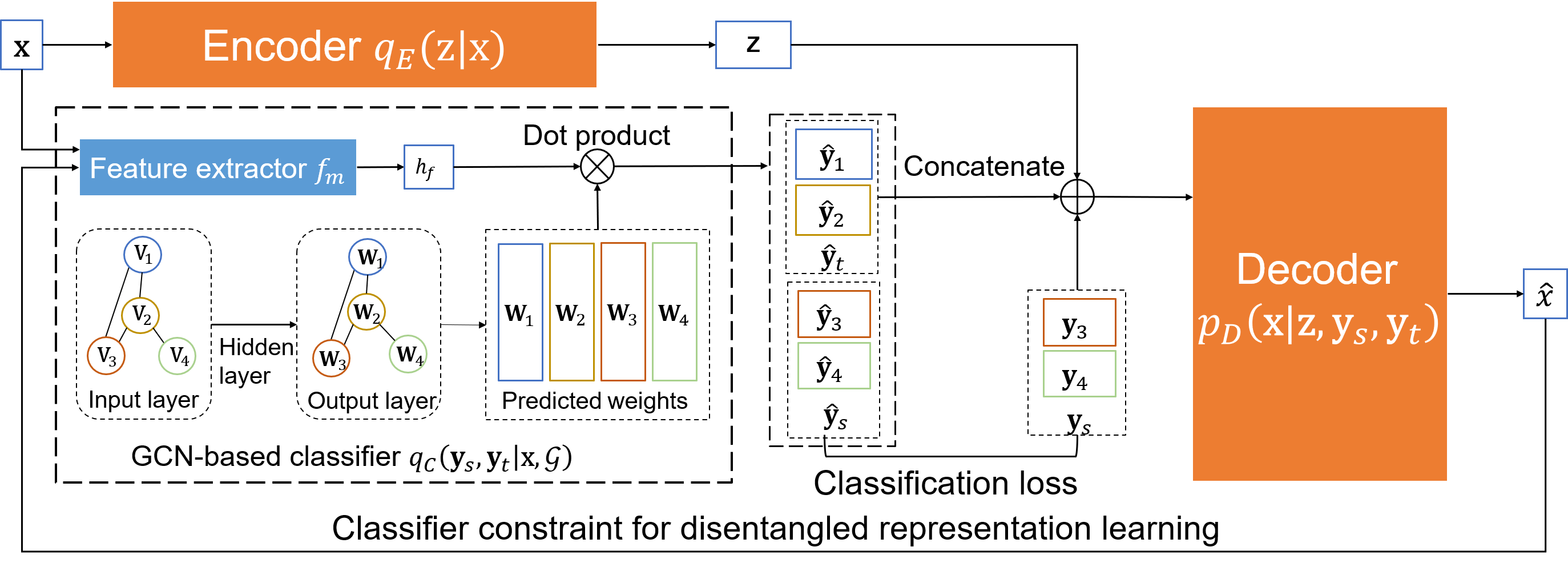}
    \vskip -1.5em
    \caption{Overall model architecture of ADDES. It shows how to process data with labels of inexact supervision classes.}
    \label{fig:Framework}
    \vskip -1em
\end{figure*}

The proposed generative framework consists of three modules: an encoder $q_E(\mathbf{z}|\mathbf{x})$, a decoder $p_D(\mathbf{x}|\mathbf{z}, \mathbf{\mathbf{y}}_s, \mathbf{\mathbf{y}}_t)$ and a GCN-based classifier $q_C(\mathbf{\mathbf{y}}_s, \mathbf{\mathbf{y}}_t|\mathbf{x}, \mathcal{G})$, which are presented in Figure \ref{fig:Framework}. The encoder is to learn a latent variable $\mathbf{z}$ disentangled with the inexact supervision label 
vector $\mathbf{y}_s$ and target label vector $\mathbf{y}_t$. The classifier utilizes the graph convolutional network to better infer $\mathbf{y}_t$ and $\mathbf{y}_s$ with the inexact supervision and unlabeled data. With the latent variable sampled from the prior $p(\mathbf{z})$ or posterior $p_E(\mathbf{z}|\mathbf{x})$, the model could synthesize a new data point $\mathbf{x}^{syn}$ corresponding to the assigned labels $(\mathbf{y}_t^{syn}, \mathbf{y}_s^{syn})$. Next, we will first introduce the probabilistic generative model for estimating data distribution followed by the deep learning framework to realize the generative model.

\begin{figure}[t]
    \centering
    \includegraphics[width=0.36\linewidth]{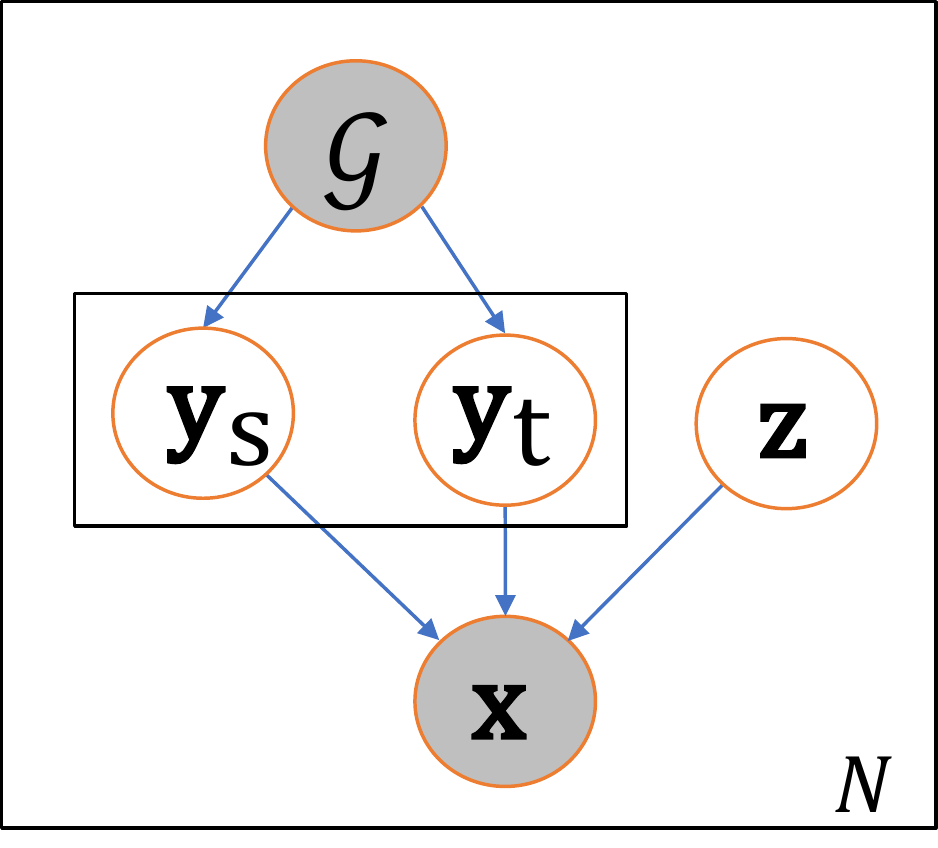}
    \vskip -1em
    \caption{The Probabilistic Graphical Model of ADDES.}
    \label{fig:PGM}
    \vspace{-3mm}
\end{figure}

\subsection{A Probabilistic Generative Model}
Our goal is to synthesize labeled data for the learning of target label predictor when only the labels of inexact supervision classes are available. To achieve this, we assume that the data is sampled from the generative process presented in Figure \ref{fig:PGM}. As shown in the figure, the data lies in a low dimension space and the latent presentation is divided into three parts: (i) $\mathbf{z}$, the latent features irrelevant with the labels; (ii) $\mathbf{y}_s$, the label vector of inexact supervision classes; (iii) $\mathbf{y}_t$, the label vector of target classes. The $\mathbf{y}_s$ and $\mathbf{y}_t$ are related with the dependency encoded by the graph $\mathcal{G}$. To generate labeled data, latent feature vector $\mathbf{z}$ is assumed to be independent with $\mathbf{y}_s$ and $\mathbf{y}_t$. With the disentangled representation, novel labeled data could be produced through varying $\mathbf{z}$, $\mathbf{y}_s$ and $\mathbf{y}_t$. Next, we give the details of the generative framework. 

\subsubsection{Modeling Data Distribution} As shown in Figure~\ref{fig:PGM}, the joint distribution $p(\mathbf{x},\mathbf{z},\mathbf{y}_s,\mathbf{y}_t|\mathcal{G})$ could be written as:
\begin{equation}
    p(\mathbf{x},\mathbf{z},\mathbf{y}_s,\mathbf{y}_t|\mathcal{G}) = p(\mathbf{x}|\mathbf{y}_s,\mathbf{y}_t,\mathbf{z})p(\mathbf{y}_s,\mathbf{y}_t|\mathcal{G})p(\mathbf{z}),
    \label{eq:distribution}
\end{equation}
where $p(\mathbf{z})$ is the prior distribution of the latent variable $\mathbf{z}$. Usually, $ p(\mathbf{z})$ is chosen as normal distribution, i.e. $p(\mathbf{z}) \sim N(\mathbf{0},\mathbf{I})$, where $\mathbf{I}$ is the identity matrix. 
For data with labels of inexact supervision classes, i.e., $(\mathbf{x},\mathbf{y}_s) \in \mathcal{D}_l$, we aim to optimize the variational lower bound (ELBO) of $\log{p(\mathbf{x}, \mathbf{y}_s|\mathcal{G})}$ as:
\begin{equation}
    \log{p(x,\mathbf{y}_s|\mathcal{G})} \geq \mathbb{E}_{q} [\log{\frac{p(\mathbf{x},\mathbf{z},\mathbf{y}_s,\mathbf{y}_t|\mathcal{G})}{q(\mathbf{y}_t,\mathbf{z}|\mathbf{y}_s,\mathbf{x},\mathcal{G})}}],
    \label{eq:ELBO}
\end{equation}
where $q(\mathbf{y}_t,\mathbf{z}|\mathbf{y}_s,\mathbf{x},\mathcal{G})$ is an auxiliary distribution to approximate $p(\mathbf{y}_t,\mathbf{z}|\mathbf{y}_s,\mathbf{x})$. To simplify the approximation process, we assume a factorized form of the auxiliary distribution:
\begin{equation}
    q(\mathbf{y}_t,\mathbf{z}|\mathbf{y}_s,\mathbf{x},\mathcal{G}) = q_C(\mathbf{y}_t|\mathbf{x},\mathcal{G})q_E(\mathbf{z}|\mathbf{x}).
\end{equation}
Then the ELBO of $\log{p(\mathbf{x}, \mathbf{y}_s|G)}$ could be re-formulated as:
\begin{equation}
    \begin{aligned}
    \log{p(\mathbf{x},\mathbf{y}_s|\mathcal{G})} & \geq \mathbb{E}_{q}[\log{p_D(\mathbf{x}|\mathbf{y}_s,\mathbf{y}_t,\mathbf{z})}] - KL[q_E(\mathbf{z}|\mathbf{x})||p(\mathbf{z})]\\
    & - KL[q_C(\mathbf{y}_t|\mathbf{x},\mathcal{G})||p(\mathbf{y}_t|\mathbf{y}_s,\mathcal{G})]\\
    & = \mathcal{L}_{G}^{l}(\mathbf{x}, \mathbf{y}_s),
    \end{aligned}
    \label{eq:ELBO_L}
\end{equation}
Similarly, for unlabeled instance $\mathbf{x} \in \mathcal{D}_u$,   we aim to optimize the variational lower bound of $\log{p(\mathbf{x}|\mathcal{G})}$:
\begin{equation}
    \begin{aligned}
    \log{p(x|\mathcal{G})} & \geq \mathbb{E}_{q} [\log{\frac{p( \mathbf{x},\mathbf{z},\mathbf{y}_s,\mathbf{y}_t|\mathcal{G})}{q(\mathbf{y}_s,\mathbf{y}_t,\mathbf{z}|\mathbf{x},\mathcal{G})}}]  \\
    & = \mathbb{E}_{q}[\log{p_D(\mathbf{x}|\mathbf{y}_s,\mathbf{y}_t,\mathbf{z})}]  - KL[q_E(\mathbf{z}|\mathbf{x})||p(\mathbf{z})]\\
    & - KL[q_C(\mathbf{y}_t, \mathbf{y}_s|x,\mathcal{G})||p(\mathbf{y}_t,\mathbf{y}_s|\mathcal{G})] \\
    & = \mathcal{L}_{G}^u(\mathbf{x}),
    \end{aligned}
    \label{eq:ELBO_U}
\end{equation}
With Eq.(\ref{eq:ELBO_L}) and Eq.(\ref{eq:ELBO_U}), the loss function on the whole training set $\mathcal{D}$ could be written as:
\begin{equation}
    \mathcal{L}_{G} = 
    \mathbb{E}_{p_l(\mathbf{x}, \mathbf{y}_s)}
    \mathcal{L}_{G}^{l}(\mathbf{x}, \mathbf{y}_s) +
    \mathbb{E}_{p_u(\mathbf{x})}
    \mathcal{L}_{G}^{u}(\mathbf{x}),
    \label{eq:VAE}
\end{equation}
where $p_l(\mathbf{x}, \mathbf{y}_s)$ denotes the distribution of $\mathcal{D}_l$, and $p_u(\mathbf{x})$ denotes the distribution of unlabeled dataset $\mathcal{D}_u$.

\subsubsection{Enforcing Disentangled Representation Learning}
The representation $\mathbf{z}$ should not contain any label information so that we can vary $\mathbf{y}_s$ and $\mathbf{y}_t$ to generate labeled data by sampling from $p(\mathbf{x}| \mathbf{z}, \mathbf{y}_s, \mathbf{y}_t)$. However, since $\mathbf{x}$ covers the information of the labels,  the latent variable $\mathbf{z}$ obtained from the encoder $q_E(\mathbf{z}|\mathbf{x})$ might correlate with $\mathbf{y}_s$ and $\mathbf{y}_t$. Thus, $\mathbf{y}_s, \mathbf{y}_t$ actually may not contribute to the generation of $\mathbf{x}$ as $\mathbf{z}$ already contains the label information. Therefore, we need to ensure $\mathbf{z}$ is independent with the class-attribute. To learn the disentangled representations, we add a constraint to enforce the data produced from the decoder 
to match the assigned labels. The objective function could be formulated as:
\begin{equation}
    \mathcal{L}_{cons} = \mathbb{E}_{p(\mathbf{y}_s,\mathbf{y}_t)p_D(\hat{\mathbf{x}}|\mathbf{y}_s,\mathbf{y}_t,\mathbf{z})}[-\log{q_C(\mathbf{y}_s,\mathbf{y}_t| \hat{\mathbf{x}}, \mathcal{G} )}],
    \label{eq:constraint}
\end{equation}
where $\hat{\bf x} = D(\mathbf{z},\mathbf{y}_s,\mathbf{y}_t)$ is the generated data from the decoder with sampled latent variable $\mathbf{z}$ and class-attribute. With this constraint, during training, we can vary $\mathbf{y}_s, \mathbf{y}_t$ for various data of desired labels. The regularizer will then check if the generated $\hat{\bf x}$ has the desired labels to enforce the involvement of $\mathbf{y}_s, \mathbf{y}_t$ in data generation.
The distribution $p(\mathbf{y}_s, \mathbf{y}_t)$ has multiple choices. When it is the prior distribution $p(\mathbf{y}_s, \mathbf{y}_t|\mathcal{G})$. This would enable the sampled data $\mathbf{x}^{syn}$ have the desired labels.
When $p(\mathbf{y}_s, \mathbf{y}_t)$ is the posterior $q_D(\mathbf{y}_t^l|\mathbf{x}^l,\mathcal{G})$ of data labeled as $\mathbf{y}_s^l$ in inexact supervision classes or posterior $q_D(\mathbf{y}_s^u,\mathbf{y}_t^u|\mathbf{x}^u,\mathcal{G})$ of unlabeled data, this constraint will assist the reconstruction of input data by providing extra semantic level supervision.

To obtain disentangled representation, $q_C(\mathbf{y}_s,\mathbf{y}_t|x,\mathcal{G})$ is used as a classifier to constrain the encoder and decoder in Eq.(\ref{eq:constraint}). This implies the predictions of the classifier are accurate.
However, the presented loss functions may be not sufficient to model the classifier well, because the predictive distribution of the classifier on $\mathcal{D}_l$ is only optimized to follow the prior distribution $p(\mathbf{y}_t|\mathbf{y}_s,\mathcal{G})$ in Eq.(\ref{eq:ELBO_L}).
And the provided labels of inexact supervision classes from the $\mathcal{D}_l$ do not contribute to model $q_C(\mathbf{y}_s,\mathbf{y}_t|\mathbf{x}, \mathcal{G})$.
This is undesirable because the distribution is used to get the label vectors of the input to generate or reconstruct the data. Thus, to better model the label predictive distribution and provide more reliable supervision for encoder and decoder, we add the loss function that explicitly utilizes the data with labels in inexact supervision classes:
\begin{equation}
    \mathcal{L}_C^{s} = \mathbb{E}_{p_l(\mathbf{x},\mathbf{y}_s)} [- \log{q_C(\mathbf{y}_s| \mathbf{x},\mathcal{G})}].
    \label{eq:C_s}
\end{equation}

\subsubsection{Final Objective Function}
Combining the variational lower bound of the generative model, the constraint to enforce the disentangled representation learning and the additional classification loss to better model the classifier, the final objective function is:
\begin{equation}
    \min_{\phi_E,\phi_C,\phi_D} \mathcal{L}_{G} + \alpha \mathcal{L}_{cons} + \beta \mathcal{L}_C^s,
    \label{eq:total}
\end{equation}
where $\alpha$ and $\beta$ are hyperparameters. And $\phi_E$, $\phi_C$, and  $\phi_D$ denote the learnable parameters  of the encoder, classifier,  and decoder.

\subsection{Deep Learning Framework of ADDES}

With the generative framework given above, we introduce the details of modeling the encoder $q_E(\mathbf{z}|\mathbf{x})$, the decoder $p_D(\mathbf{x} | \mathbf{y}_s, \mathbf{y}_t, \mathbf{z})$, and the classifier $q_C(\mathbf{y}_t, \mathbf{y}_s | x, \mathcal{G})$ now.

\subsubsection{Encoder and Decoder} For many applications such as images and text, both $q_E(\mathbf{z}|\mathbf{x})$ and $p_D(\mathbf{x} | \mathbf{y}_s, \mathbf{y}_t, \mathbf{z})$ could be very complex distributions. Following VAE~\cite{kingma2013autoencoding}, we use neural network and reparameterization trick to model $q_E(\mathbf{z}|\mathbf{x})$ and $p_D(\mathbf{x} | \mathbf{\mathbf{y}}_s, \mathbf{y}_t, \mathbf{z})$, which are shown to be able to approximate complex distributions under mild conditions. Specifically, we assume the encoder $q_E(\mathbf{z}|\mathbf{x})$ follows Gaussian distribution with the mean and variance as the output of a neural network:
\begin{equation}
    q_E(\mathbf{z}|\mathbf{x}) = N(\mathbf{z}; \boldsymbol{\mu}_z, \boldsymbol{\sigma}_z^2 \mathbf{I}), \quad \boldsymbol{\mu}_z, \boldsymbol{\sigma}_z = E(\mathbf{x}) 
\end{equation}
where $E(\cdot)$ is the neural network which takes $\mathbf{x}$ as input and output the mean $\boldsymbol{\mu}_z$ and standard deviation $\boldsymbol{\sigma}_z$. Then $\mathbf{z}$ can be sampled as $\mathbf{z} = \boldsymbol{\mu}_z + \boldsymbol{\sigma}_z \odot \boldsymbol{\epsilon}$, where $\boldsymbol{\epsilon}$ is sampled from a normal distribution. Similarly, we assume the decoder $p_D(\mathbf{x} | \mathbf{y}_s, \mathbf{y}_t, \mathbf{z})$ follows Gaussian distribution with the mean and variance as the output of a deep neural network:
\begin{equation}
    p_D(x | \mathbf{y}_s, \mathbf{y}_t, z) = N(x; \boldsymbol{\mu}_x, \boldsymbol{\sigma}_x^2 \mathbf{I}), \quad \boldsymbol{\mu}_x, \boldsymbol{\sigma}_x = D(\mathbf{y}_s, \mathbf{y}_t, \mathbf{z}) 
\end{equation}
where $D(\cdot)$ is the neural network which takes $(\mathbf{y}_s, \mathbf{y}_t, \mathbf{z})$ as input and output the mean $\boldsymbol{\mu}_x$ and standard deviation $\boldsymbol{\sigma}_x$. The structure of the $E(\cdot)$ and $D(\cdot)$ can be chosen based on the domain we are working on. For example, for image datasets, deep convolutional neutral networks could be applied. For text datasets, sequence to sequence models are good candidates. 

\subsubsection{GCN-based Classifier}
Since only the inexact supervision is available, we rely on the dependency between the labels of inexact supervision classes and target classes to infer the target labels, and the dependency is encoded in the graph $\mathcal{G}$. Graph neural networks have been demonstrated to be very effective in capturing the relationship between nodes in a graph. Therefore, to model $q_C(\mathbf{y}_t, \mathbf{y}_s|\mathbf{x}, \mathcal{G})$, we adopt Graph Convolutional Networks (GCN) for $\mathcal{G}$ and propose a GCN-based classifier. 
Figure~\ref{fig:Framework} gives an illustration of the GCN-based classifier, which consists of two parts, i.e., a feature extractor and a GCN module. The basic idea is to learn representations of classes from $\mathcal{G}$ using GCN and the features of $\mathbf{x}$ using the feature extractor, then conduct label prediction based on these representations. 

\textbf{Feature extraction:} 
To facilitate the classification by the GCN module, a low dimension representation of $\mathbf{x}$ is required. 
One way is to use  the latent feature $\mathbf{z}$ from the encoder $E(\mathbf{x})$. However, since the encoder is expected to learn $\mathbf{z}$ that has no semantic information about the labels, directly using $\mathbf{z}$ cannot help predict labels. Thus, another feature extractor is required. The model architecture is quite flexible. For images, a CNN model such as AlexNet \cite{krizhevsky2012imagenet}, VGG \cite{simonyan2014very} and ResNet \cite{he2016deep} can be  feature extractor. For text, LSTM \cite{hochreiter1997long}, GRU \cite{cho2014learning}, CNN \cite{kim2014convolutional} and transformer \cite{vaswani2017attention} are all potential models. With the feature extractor model $f_{m}$, we could attain the representation of input $\mathbf{x}$ as
\begin{equation}
    \mathbf{h}_f = f_{m}(\mathbf{x})\in \mathbb{R}^{F},
\end{equation}
where $F$ denotes the dimension of the extracted feature. 

\textbf{Synthesize classifiers with GCN:} 
The GCN is to generate the parameters of classifiers for both inexact supervision classes and target classes. Each node of the graph corresponds to a class in the whole class set $\mathcal{W}$. Thus, the number of nodes is $n = |\mathcal{W}|$. The adjacency matrix of the graph is $ \mathbf{A} \in \mathbb{R}^{n \times n}$. And $\mathbf{A}_{ij}$ indicates the how strong the correlation between the labels of classes $l_i$ and $l_j$ is.
The GCN-layer updates the node features $\mathbf{H} \in \mathbb{R}^{n \times d}$ by aggregating the information from the neighbors, where $d$ represents the dimension of the node features.  The process can be written as:
\begin{equation}
    \mathbf{H}^{l+1} = f (\tilde{\mathbf{D}}^{-1}\tilde{\mathbf{A}}\mathbf{H}^{l}\mathbf{W}^{l} ),
\end{equation}
where $\tilde{\mathbf{A}} = \mathbf{A} + \mathbf{I} $ and $\tilde{\mathbf{D}}$ is the degree matrix of $\tilde{\mathbf{A}}$. $f$ denotes the nonlinear active function. $\mathbf{W}^l \in \mathbb{R}^{d^{l} \times d^{l+1}}$ is the weights of the $l$-th layer, where $d^{l}$ is the dimension of the latent feature in the $l$-th layer. The input of the first layer $\mathbf{V}$ could be word embeddings or one-hot embeddings of the classes. The output of the last GCN layer is $\mathbf{W} \in \mathbb{R}^{n \times F}$, which corresponds to classifier weights of the 
classes. The predicted scores of all the classes including inexact supervision classes and target classes could be obtained by:
\begin{equation}
    \begin{bmatrix} \hat{\bf y}_{s} \\ \hat{\bf y}_{t}\end{bmatrix} = \sigma ( \begin{bmatrix}\mathbf{W}_{s}\\ \mathbf{W}_{t}\end{bmatrix} \mathbf{h}_f),
\end{equation}
where $\mathbf{W}_s$ and $\mathbf{W}_t$ indicate the synthesized classifier weights of the inexact supervision classes and target classes. $\hat{\bf y}_{s} \in \mathbb{R}^{|\mathcal{S}|}$ with the $i$-th element denoting the probability that the label of $i$-th inexact supervision class being 1. Similarly, $\hat{\bf y}_{t} \in \mathbb{R}^{|\mathcal{T}|}$ with the $j$-th element denoting the probability that the $j$-th target label being 1. 
With the parameter sharing and explicit utilization of graph structure, the inexact supervision could be propagated to the target classes to obtain reasonable classifiers.

\begin{algorithm}[t] 
\caption{ Training Algorithm of ADDES.} 
\label{alg:Framwork} 
\begin{algorithmic}[1]
\REQUIRE
$\mathcal{D}_l=\{(\mathbf{x}^l, \mathbf{y}_s^l)\}$, $\mathcal{D}_u=\{\mathbf{x}^u\}$, $\mathbf{A}$, $\mathbf{V}$, $\alpha$ and $\beta$.
\ENSURE $q_E(\mathbf{z}|\mathbf{x})$, $p_D(\mathbf{x} | \mathbf{y}_s, \mathbf{y}_t, \mathbf{z})$, and $q_C(\mathbf{y}_t, \mathbf{y}_s | \mathbf{x}, \mathcal{G})$

\STATE Initialize the GCN-based classifier by minimizing Eq.(\ref{eq:C_s}) and the third term of Eq.(\ref{eq:ELBO_L}).
\STATE Initialize the encoder and decoder by minimizing Eq.(\ref{eq:total}) with the parameters of the classifier frozen.
\REPEAT 
\STATE 
Randomly sample $\{\mathbf{x}^l_i\}_{i=1}^N$ from $\mathcal{D}_l$ and $\{\mathbf{x}^u_i\}_{i=1}^N$ from $\mathcal{D}_u$;\\
\STATE Optimized the encoder parameters $\phi_E$, decoder parameters $\phi_D$ and classifier parameters $\phi_C$ by Eq.(\ref{eq:total}). 

\UNTIL convergence
\RETURN $q_E(\mathbf{z}|\mathbf{x})$, $p_D(\mathbf{x} | \mathbf{y}_s, \mathbf{y}_t, z)$, and $q_C(\mathbf{y}_t, \mathbf{y}_s | \mathbf{x}, \mathcal{G})$
\end{algorithmic}
\end{algorithm}
\subsection{Training Algorithm}
The overall training algorithm of ADDES is given in Algorithm \ref{alg:Framwork}. Firstly, before jointly training the encoder, decoder, and classifier, these modules are separatly pretrained to have good initialization parameters. More specifically, the GCN classifier is prioritize to be optimized. Then with the classifier's parameters fixed, the encoder and decoder are pretrained with Eq.(\ref{eq:total}). Secondly, to make the gradients able to backpropagate from the decoder to the classifier, we directly 
input the soft labels to the decoder. 

\section{Experiments} \label{sec:experiments}
In this section, we conduct a series of experiments to validate the effectiveness of our proposed framework. They are designed to answer the following research questions: 
\begin{itemize}[leftmargin=*]
    \item \textbf{RQ1} Could the proposed generative model synthesize useful labeled data as data augmentation for target label prediction?
    \item \textbf{RQ2} Could our proposed method bring benefits to different scenarios whose training data varies in types and sizes?      
    \item \textbf{RQ3} Does the utilization of graph structure of labels promote the generative model learning? If it works, is it sensitive to the graph construction method?
\end{itemize}
\subsection{Datasets}
We conduct experiments on two publicly available datasets, including a text dataset StackOverflow and an image dataset MJSynth. 

\textbf{StackOverflow\footnote{https://www.kaggle.com/stackoverflow/StackOverflow}}: It contains texts of 10\% of questions and answers from the Stack Overflow programming Q\&A website. The majority of the questions have multiple tags. 
After filtering out rare tags, we obtain a tag set of size 25. Each question is labeled with 1.9 tags on average. 
To demonstrate ADDES could synthesize useful labeled data to facilitate the classification for various target classess, two sets from the 25 tags are set as target classes and sequently educe two datasets.
Specifically, the target class sets are $\mathcal{T}_1 = \{ \textit{javascript}\}$ and $\mathcal{T}_2=\{ \textit{javascript, C\#, php, java}\}$, which refer to as \textbf{StackOverflow-1} and \textbf{StackOverflow-2}, respectively.
For both datasets, the size of $\mathcal{D}_l$ is 2k.
The unlabeled set $\mathcal{D}_u$ contains 30k questions. For the test set, we randomly sample 30k questions.

\textbf{MJSynth}~\cite{Jaderberg14c}:  It is used for natural scene text recognition. Each image in MJSynth contains a word extracted from the text corpus. The images are produced in a sophisticatedly designed pipeline to emulate the text in the natural scene. 
In the MJSynth dataset, the character label indicates if a certain letter is in the image or not. We filter out the images that contain characters other than the lower case alphabet, which makes the number of classes to 26. The number of character labels per image is 6 on average.
Similar to StackOverflow, two target class sets,  $\mathcal{T}_1 = \{\textit{e}\}$ and $\mathcal{T}_2=\{\textit{a},\textit{c},\textit{e}\}$ are selected to build new datasets named as \textbf{MJSynth-1} and \textbf{MJSynth-2}. For both datasets, we sample 6k and 24k images as $\mathcal{D}_l$ and  $\mathcal{D}_u$. Another 10k images are sampled as test sets.

\subsection{Baselines}
We compare our method with the following state-of-the-art baselines from the supervised classification, semi-supervised learning, multi-label learning and zero-shot learning:
\begin{itemize}[leftmargin=*]
    \item \textbf{text-CNN}~\cite{kim2014convolutional}: Convolutional filters with different kernel sizes are applied to get the features for text classification. 
    \item \textbf{GRU}~\cite{cho2014learning}: It utilizes  GRU cell to extract the text features.
    \item \textbf{SDANN}~\cite{Jaderberg14c}: This is a network with five convolutional layers to process MJSynth for natural text recognition.
    \item \textbf{Semi-CNN}~\cite{grandvalet2005semi}: Semi-CNN utilizes unlabeled data $\mathcal{D}_u$ by adding an entropy regularization term to train the classifier. 
    \item \textbf{ML-GCN}~\cite{chen_multi-label_2019}: A state-of-the-art method applies the GCN to model the label relations for multi-label classification.
    \item \textbf{Zero-Shot}~\cite{Wang_2018_CVPR}: This is a state-of-the-art method for zero-shot learning. It transfers the knowledge learned from seen labels prediction by employing the GCN to predict the weights of classifiers for target label prediction. Thus, only the instance and labels of inexact supervision classes are required.
\end{itemize}
\vspace{-1mm}
Aside from ML-GCN and Zero-Shot, these baselines assume labels of different classes are independent with each other.
Moreover, all these methods require labels of target classes except Zero-Shot. Therefore, we develop ways to attain estimated target labels $\hat{\bf y}_t$ through inexact supervision labels $\mathbf{y}_s$ in the following subsections.

\subsection{Experiments on the Text Datasets}

\subsubsection{Graph Construction}
The edges between the inexact supervision classes $l_s \in \mathcal{S}$ are built based on the conditional probability $P(l_i|l_j)$, which is the probability of an instance belonging to class $l_i$ when it is known to be in class $l_j$. 
According to~\cite{chen_multi-label_2019}, we count the occurrence of label pairs from $\mathcal{D}_l$ and obtain the matrix $\mathbf{M} \in \mathbb{R}^{|\mathcal{S}| \times |\mathcal{S}|}$, where $\mathbf{M}_{ij}$ represents the count of instances labeled as 1 in both $l_i$ and $l_j$. Then, the conditional probability is:
\begin{equation}
    \mathbf{A}_{ij}=P(l_i|l_j) = \frac{\mathbf{M}_{ij}}{N_j}, \quad l_i, l_j \in \mathcal{S}
    \label{eq:correlation}
\end{equation}
where $N_j$ denotes the count of instances which belong to $l_j$ in the dataset. With Eq. \ref{eq:correlation}, the weights of the edges linking the inexact supervision classes could be obtained.
However, due to the lacking of annotations of target labels in the training set, we are unable to build the edges between the target classes and inexact supervision classes by Eq.(\ref{eq:correlation}). Introducing the prior knowledge to the graph construction could solve this problem. For instance, based on the primary programming knowledge, we could add undirected edges between the target class \texttt{javascript} and 
inexact supervision class \texttt{ajax}. With the $t$-th target class denoted as $l_t$ and its manually assigned related class set denoted as $\mathcal{R}_t$, the process of linking the target and inexact supervision classes can be formulated as:
\begin{equation}
    \mathbf{A}_{ts} = \mathbf{A}_{st}=\begin{cases}
                    1& ,\text{if }  l_s \in \mathcal{R}_t\\
                    0& ,\text{else} 
                    \end{cases},
    \label{eq:target_link}
\end{equation}
where $l_s \in \mathcal{S}$ is the $s$-th inexact supervision class.
The constructed graph could also be used to estimate the conditional probability $p(\mathbf{y}_t|\mathbf{y}_s, \mathcal{G})$ which is required in Eq.(\ref{eq:ELBO_L}). The estimation formula is:
\begin{equation}
    P(\mathbf{y}_t(i)=1|\mathbf{y}_s,\mathcal{G}) = \begin{cases} 1 & , \text{if $\exists l_j \in \mathcal{S}, \mathbf{y}_s(j)=1 \cap \mathbf{A}_{ij}=1$ }\\
                               0 &, \text{else}
                \end{cases},
    \label{eq:noisy_labels}
\end{equation}
where $\mathbf{y}_t(i)$ denotes the label of the target class $l_i \in \mathcal{T}$, and $\mathbf{y}_s(j)$ means the label of inexact supervision class $l_j \in \mathcal{S}$.


\begin{table*}[t!]
    \small
    \centering
    \caption{The comparisons with the baselines and models trained with text dataset augmented with synthetic data.}
    \vskip -1.5em
    \begin{tabularx}{0.96\linewidth}{|p{0.12 \linewidth}|p{0.04 \linewidth}|C|C|C|C|C|C|C|}
    \hline
         Datasets & Metric & text-CNN & GRU & ML-GCN & Semi-CNN & Zero-shot & AugCNN &AugGCN  \\
    \hline
    \hline
    \multirow{2}{*}{StackOverflow-1}  
    &  mAP & 0.528 $\pm 0.004$  & 0.516 $\pm 0.007$ & 0.565 $\pm 0.008$ & 0.549 $\pm 0.004$ & 0.541 $\pm 0.003$ & 0.609 $\pm 0.003$ & \textbf{0.629} $\pm \mathbf{0.003}$\\
    \cline{2-9}
    &AUC & 0.773 $\pm 0.004$ & 0.763 $\pm 0.004$ & 0.788 $\pm 0.003$ & 0.782 $\pm 0.002$ & 0.781 $\pm 0.002$ & 0.830 $\pm 0.001$ & \textbf{0.832} $\pm \mathbf{0.003}$\\
    \hline
    \hline
     \multirow{2}{*}{StackOverflow-2}    
     &   mAP         & 0.350 $\pm 0.003$ & 0.342 $\pm 0.008$ & 0.379 $\pm 0.005$ & 0.376 $\pm 0.003$  & 0.369 $\pm 0.002$ & 0.400 $\pm 0.003$ & \textbf{0.412} $\pm \mathbf{0.004}$\\
    \cline{2-9}
     &   AUC & 0.666 $\pm 0.003$  & 0.646 $\pm 0.006$ & 0.685 $\pm 0.003$ & 0.663 $\pm 0.002$ & 0.659 $\pm 0.005$ & 0.686 $\pm 0.003$ & \textbf{0.703} $\pm \mathbf{0.003}$\\
    \hline
    \end{tabularx}
    \label{tab:text_result}
\end{table*}

\subsubsection{Implementation Details}

The encoder of the ADDES for text generation is based on the bi-directional GRU with the hidden dimension set as 150. The mean and variance of the latent variable could be obtained from the hidden states of the GRU cell. For decoder, we adopt a global attention mechanism  \cite{luong2015effective} to facilitate  focusing on critical parts of the input sequence. 
Similarly, bi-directional GRU is applied to extract features for the GCN-based classifier. The GCN module of the classifier has two layers. One-hot embeddings are used as node attributes. The hyperparameters of ADDES are: $\alpha=0.1, \beta=0.1$.


\begin{figure}[t]
\centering
\begin{subfigure}{0.47\columnwidth}
    \centering
    \includegraphics[width=\linewidth]{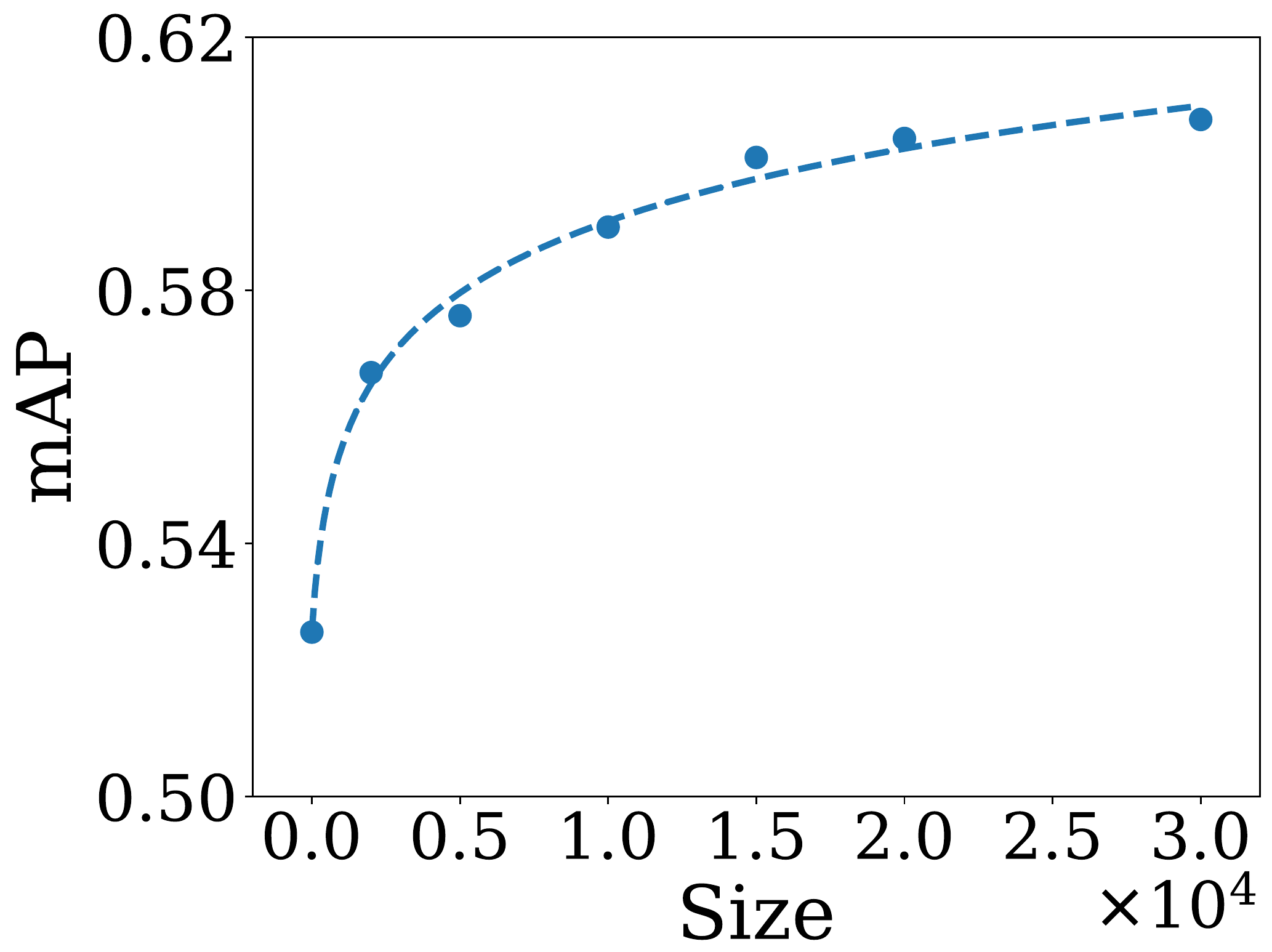} 
    \vspace{-7mm}
    \caption{StackOverflow-1}
    \label{fig:syn_text_AP}
\end{subfigure}
\begin{subfigure}{0.47\columnwidth}
    \centering
    \includegraphics[width=\linewidth]{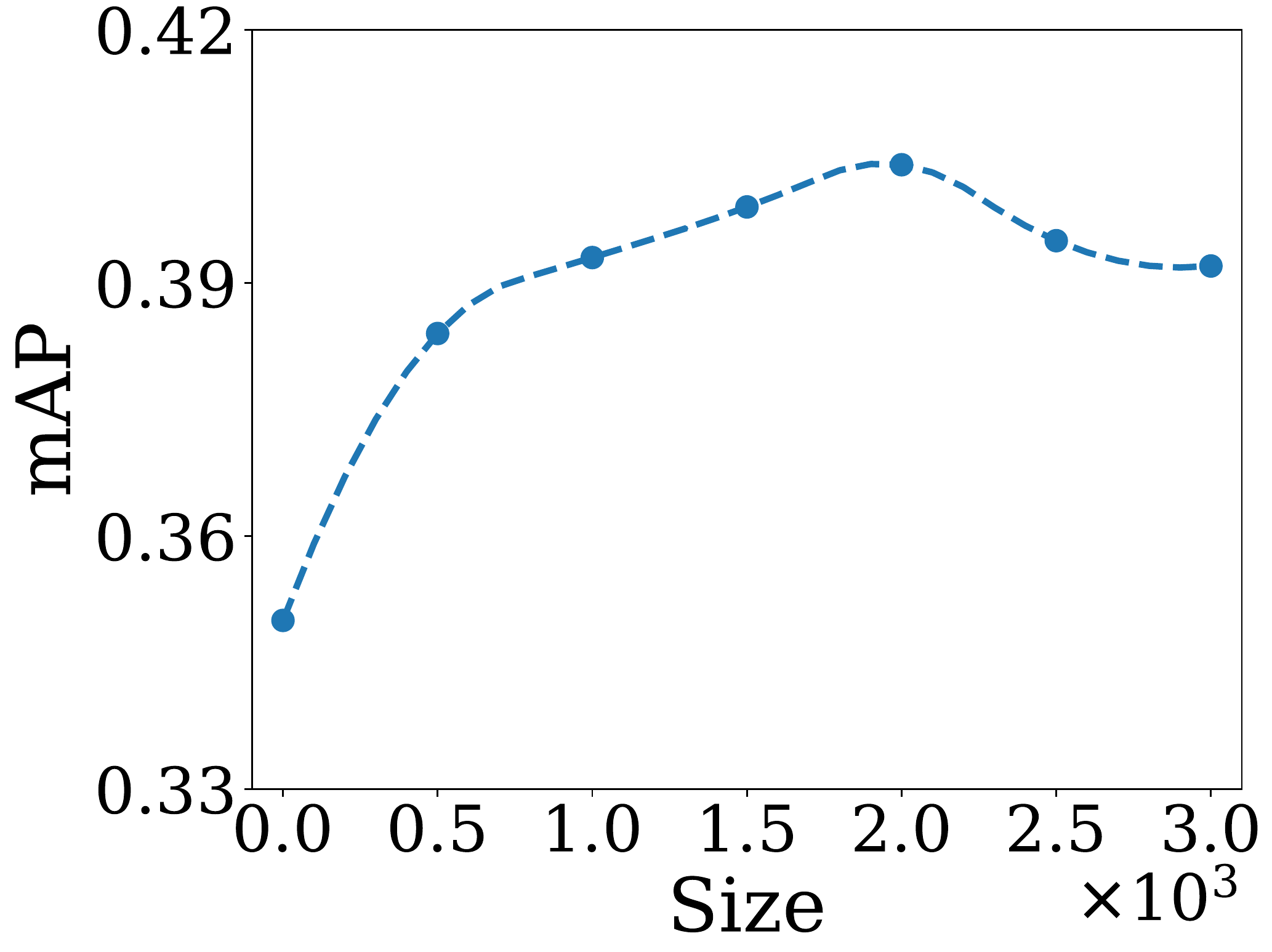} 
    \vspace{-7mm}
    \caption{StackOverflow-2}
    \label{fig:syn_text_mAP}
\end{subfigure}
\vskip -1em
\caption{The performance of the text-CNN w.r.t the numbers of synthetic data from ADDES added to the training set.}
\vskip -1em
\label{fig:Syn_text_curve}
\end{figure}

\begin{table}[t]
    \centering
    \small
    \vskip -1em
    \caption{The synthetic labeled data for StackOveflow-1.}
    \vskip -1.5em
    \begin{tabularx}{0.98\columnwidth}{Xp{0.65\columnwidth}}
    \toprule
         Input Labels &  Generated Text\\
         \midrule
         javascript & Angular 2 Routing in plain Javascript\\
         javascript & Different CSS depending on month and year\\
         javascript, jquery  & JQuery replace on Click in h \\
        \midrule
        C++ &  C++ : find in set of pointers \\
        java, json & query elasticsearch with java with JSON\\
        C++, Android & Display image created by OpenCV on Android\\
        \bottomrule
    \end{tabularx}
    \vskip -1em
    \label{tab:synthetic_text}
\end{table}

\subsubsection{Experimental Results}
To answer \textbf{RQ1}, we synthesize a set of labeled data $\mathcal{D}_s=\{(x^{syn_i},\mathbf{y}_s^{syn_i},\mathbf{y}_t^{syn_i})\}^{n_s}_{i=1}$ for data augmentation, where $n_s$ is the size of synthetic dataset. We also supplement the instance containing inexact supervision, i.e., $(x, \mathbf{y}_s) \in \mathcal{D}_l$ with estimated target label $ \hat{\mathbf{y}}_t \in \{0, 1\}^{|\mathcal{T}|}$ through Eq.(\ref{eq:noisy_labels}) to 
build estimated labeled dataset $\mathcal{D}_e$.
Then the performance of the classifiers trained with $\mathcal{D}_s \cup \mathcal{D}_e$ could show whether the synthetic data could bring benefit to the classifiers for target label prediction. The performance of the models is evaluated by two metrics: mean average precision (mAP) and the average area under ROC curve (AUC).

\textbf{Impacts of the size of $\mathcal{D}_s$}: 
It has been reported that the number of synthetic data added into the training set could strongly affect the performance of supervised learning \cite{shin2018medical}. Therefore, we investigate the performance of the classifier whose training set is enlarged with the synthetic dataset $\mathcal{D}_s$ in different sizes. Here, different numbers of synthetic data mixed with 2k estimated labeled data 
are applied to train the text-CNN model. The 
results are shown in Figure \ref{fig:Syn_text_curve}. From Figure \ref{fig:syn_text_AP}, it is observable that the performance improves up to saturation as we add more synthetic data to the training set in StackOverflow-1. As results of StackOverflow-2 shown in Figure \ref{fig:syn_text_mAP},  we could find the gain brought by synthetic labeled data $\mathcal{D}_s$ will firstly increase and then decrease as the size of $\mathcal{D}_s$ increases. This is because there are more target classes in StackOverflow-2, which makes it more challenging to generate high-quality labeled data. For both datasets, compared with the models trained without synthetic data, i.e., $|\mathcal{D}_s|=0$, the models trained with augmented data consistently perform better. Therefore, the generated labeled text is useful as data augmentation for target label prediction. 

\begin{table*}[t]
    \small
    \centering
    \caption{The results of the baselines and the classifiers trained with augmented data on the image datasets.}
    \vskip -1.5em
    \begin{tabularx}{0.96\linewidth}{|C|C|C|C|C|C|C|C|}
    \hline
         Dataset & Metric & SDANN & ML-GCN & Semi-CNN & Zero-shot & AugCNN &AugGCN  \\
    \hline
    \hline
        \multirow{2}{*}{MJSynth-1}&mAP & 0.767 $\pm 0.004$ &  0.788 $\pm 0.005$ & 0.776 $\pm 0.003$ & 0.700 $\pm 0.001$ & 0.810 $\pm 0.004$ & \textbf{0.816} $\pm \mathbf{0.002}$\\
    \cline{2-8}
         & AUC & 0.640 $\pm 0.005$ & 0.665 $\pm 0.006$ & 0.650 $\pm 0.002$ & 0.556 $\pm 0.003$ & 0.696 $\pm 0.005$ & \textbf{0.701} $\pm \mathbf{0.003}$ \\
    \hline
    \hline
        \multirow{2}{*}{MJSynth-2}
        & mAP & 0.555 $\pm 0.005$ & 0.572 $\pm 0.005$ & 0.562 $\pm 0.001$ & 0.539 $\pm 0.006$ &  0.584 $\pm 0.005$ & \textbf{0.606} $\pm \mathbf{0.004}$\\
    \cline{2-8}
        & AUC & 0.584 $\pm 0.004$ & 0.606 $\pm 0.007$ & 0.593 $\pm 0.002$ & 0.567 $\pm 0.005$ & 0.620 $\pm 0.004$ & \textbf{0.640} $\pm \mathbf{0.005}$\\
    \hline
    \end{tabularx}

    \vskip -1em
    \label{tab:image_result}
\end{table*}

\textbf{Comparisons with baselines}: 
We evaluate the benefits brought by synthetic data to different models , i.e., text-CNN and ML-GCN, and compare them with the baselines. More specifically, models with the same structure as text-CNN and ML-GCN are trained by adding an optimal size of $\mathcal{D}_s$ to the original estimated labeled dataset $\mathcal{D}_e$, which refer to as \textbf{AugCNN} and \textbf{AugGCN}. According to Figure \ref{fig:Syn_text_curve}, the size of $\mathcal{D}_s$ is set as 30k and 2k in 
StackOverflow-1 and StackOverflow-2, respectively.
The average results and standard deviation of five runs are presented in Table \ref{tab:text_result}.
We could have the following observations: (i) AugCNN is better than the state-of-the-art multi-label classification method ML-GCN and the semi-supervised learning approach Semi-CNN; (ii) AugCNN and AugGCN outperform text-CNN and ML-GCN and other baselines with a large margin, which indicates the synthetic data could be helpful for various models. These observations confirm that the proposed model could generate high-quality labeled data when only inexact supervision is available in text datasets.

\textbf{Visualization of the synthetic text}:  Some samples of generated labeled data for StackOverflow-1 whose target class set is $\{\textit{javascript}\}$ are reported in Table \ref{tab:synthetic_text}. It shows that we could produce realistic text which contains the target label \texttt{javascript}. Furthermore, the model could also generate the text with multiple labels.

\subsection{Experiments on the Image Datasets}

\subsubsection{Graph Construction} We aim to assign links between the labels that often occur together. As each image in the datasets contains a word, the co-occurrence probability of labels should be the same as the probability that two letters appear together in a word. Therefore, we use a common word corpus YAWL\footnote{http://freshmeat.sourceforge.net/projects/yawl/} to calculate the correlation matrix by Eq.(\ref{eq:correlation})
Moreover, the estimation of the conditional probability $p(\mathbf{y}_t|\mathbf{y}_s,\mathcal{G})$ could be attained through the YAML corpus. Specially,  We train a random forest model on YAML to build the estimated labeled dataset $\mathcal{D}_e$. It predicts the target label vector $\mathbf{y}_t$ based on the inexact supervision label vector $\hat{\mathbf{y}}_s$ of $\mathcal{D}_l$.

\subsubsection{Implementation Details}
The encoder is composed of 5 convolutional layers which contain 64, 128, 256, 512, and 512 filters with the kernel size and stride set as 4 and 2. The mean and variance of the latent variable are obtained from the output of the global max pooling layer of the encoder. The structure of the decoder is symmetrical to the encoder. Transpose convolution with stride 2 is used to upsample the feature map in the decoder. For the GCN-based classifier, the feature extractor has the same structure as the encoder. And there is one hidden layer with the filer size set as 128 in the classifier. The hyperparameters are set as: $\alpha=100, \beta=0.1$.

\subsubsection{Experimental Results} 
The proposed framework could also generate useful labeled data for data augmentation in image classification. 
Similar to the text datasets, we investigate impacts of size of the synthetic labeled data $\mathcal{D}_s$ to the models utilizing synthetic data. Then, we compare the performance of the baselines and the models training with augmented data to demonstrate the generated labeled data could facilitate the learning of image classifier.

\begin{figure}[t]
\centering
\vskip -1em
\begin{subfigure}{0.49\columnwidth}
    \centering
    \includegraphics[width=\linewidth]{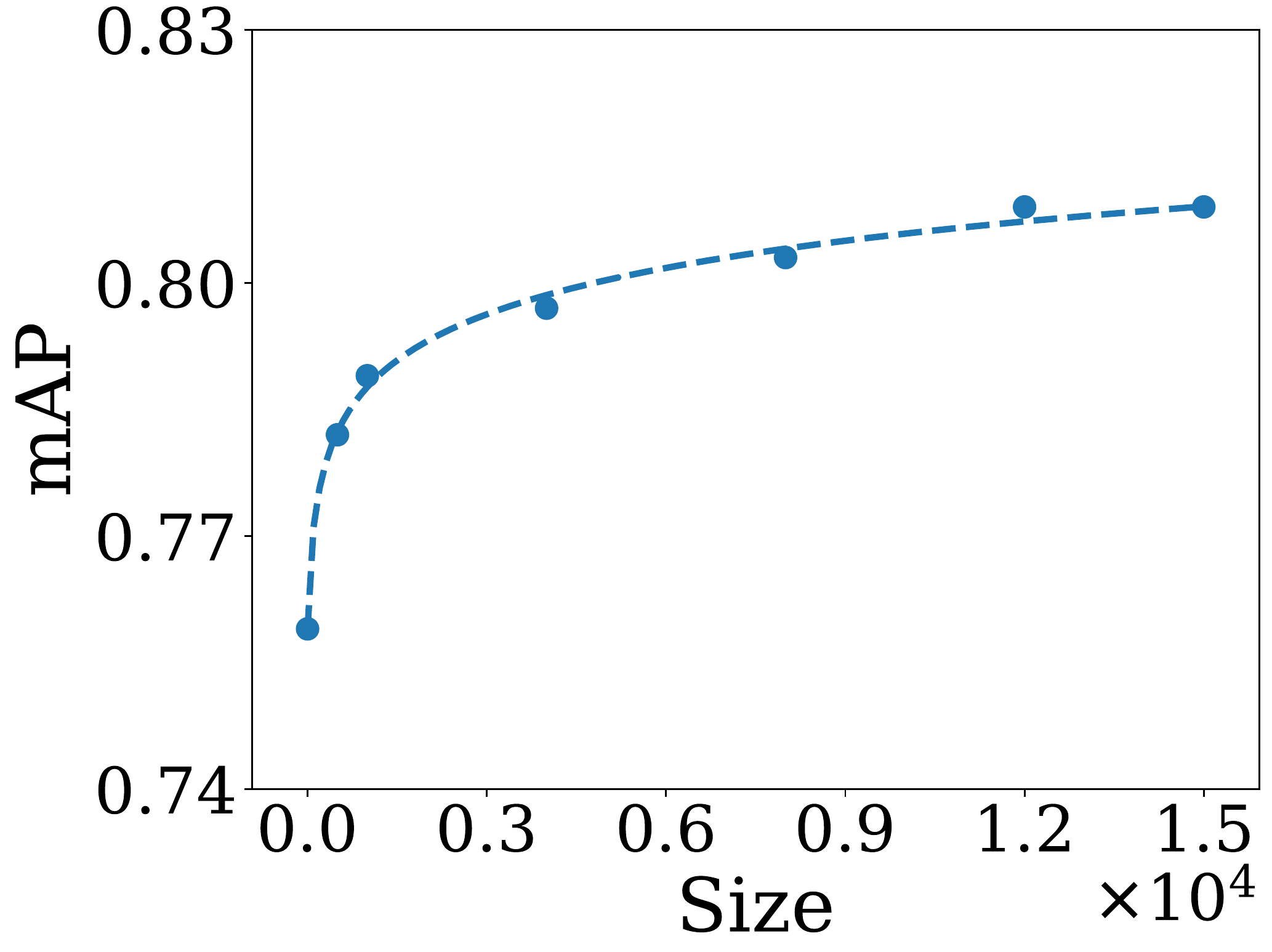} 
    \vspace{-7mm}
    \caption{MJSynth-1}
    \label{fig:syn_img_AP}
\end{subfigure}
\begin{subfigure}{0.49\columnwidth}
    \centering
    \includegraphics[width=\linewidth]{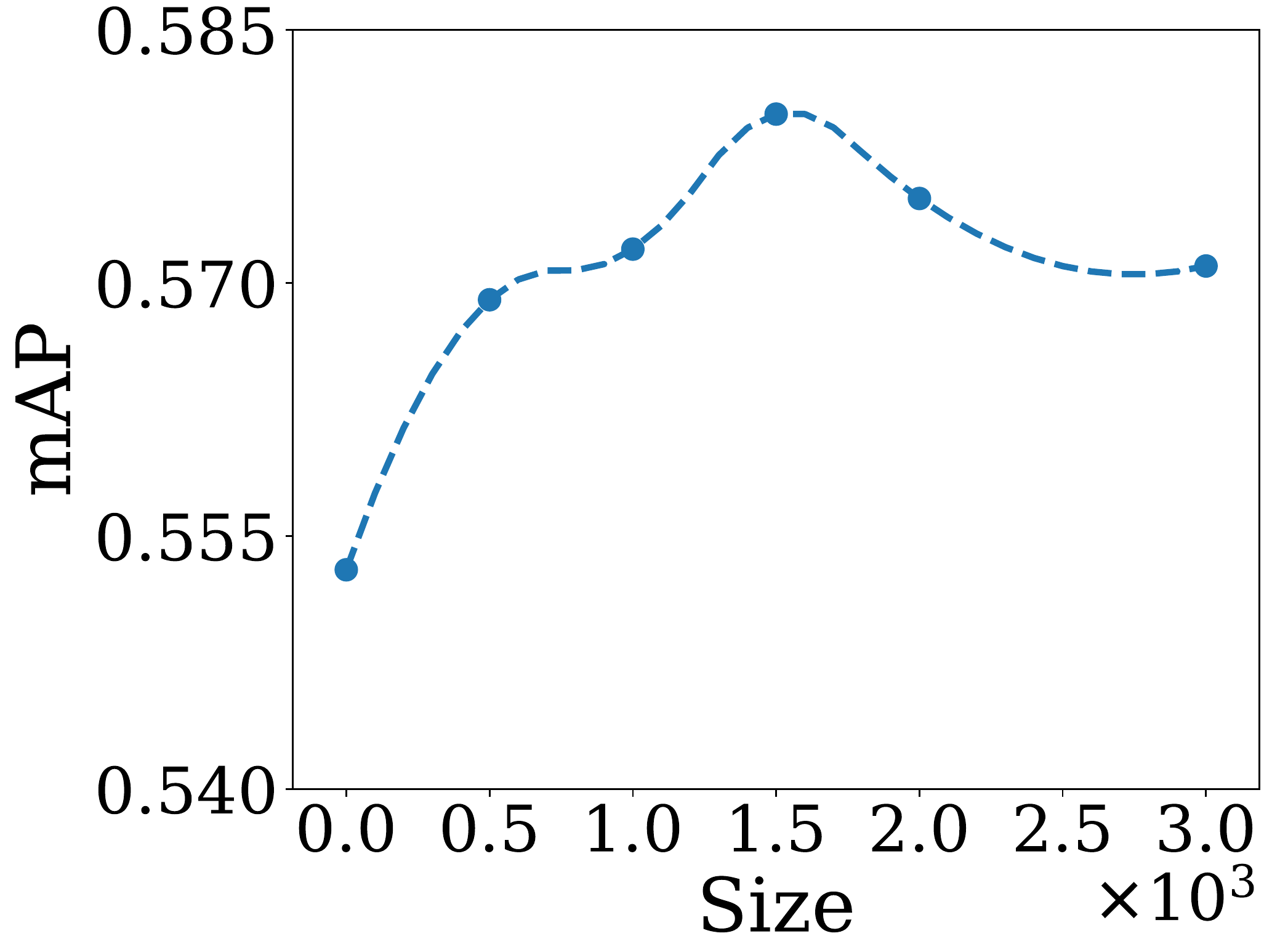} 
    \vspace{-7mm}
    \caption{MJSynth-2}
    \label{fig:syn_img_AUC}
\end{subfigure}
\vskip -1em
\caption{The performance of the SDANN w.r.t the numbers of synthetic data from ADDES added to the training set.}
\label{fig:Syn_img_curve}
\end{figure}

\textbf{Impacts of size of $\mathcal{D}_s$}: The Figure \ref{fig:Syn_img_curve} shows the trend of the performance of the SDANN with the increase of augmented images. 
We could observe that the synthetic data could improve the performance of the model for both image datasets. The trend of the curves is in line with the results of text datasets. The evident improvements after introducing a reasonable number of synthetic images demonstrate the validity of the generated labeled data. 

\begin{figure}[t]
    \centering
    \includegraphics[width=0.80\columnwidth]{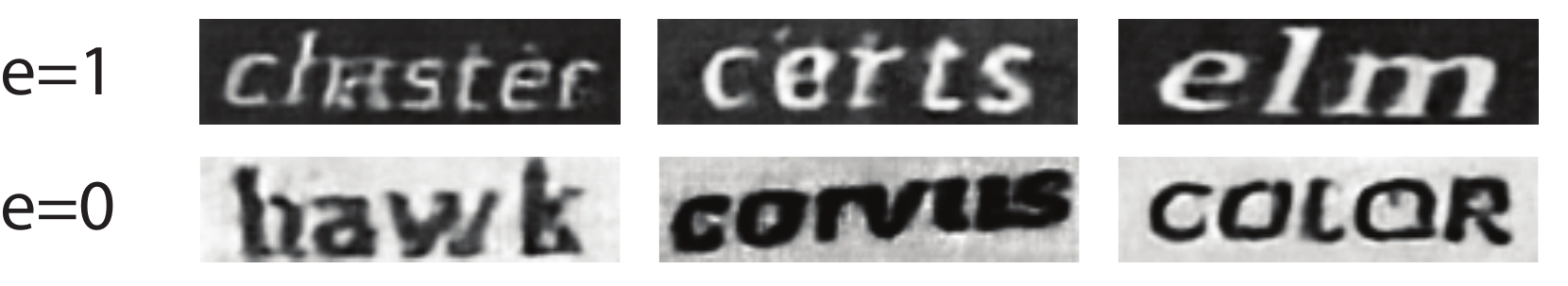}
    \vspace{-1em}
    \caption{The synthetic images of ADDES on MJSynth-1}
    \label{fig:vis_image}
    \vskip -1em
\end{figure}

\textbf{Comparisons with baselines}: We train SDANN and ML-GCN with the augmented dataset $\mathcal{D}_e \cup \mathcal{D}_s$ to get \textbf{AugCNN} and \textbf{AugGCN} and find significant improvements compared with the baselines.
From Table \ref{tab:image_result}, we could have similar observations in image datasets:
(i) AugCNN outperforms SDANN and Semi-CNN with a large margin and even performs better than the sophisticatedly designed model ML-GCN, which shows our generative model utilizes unlabeled data and label relations well; (ii) AugGCN achieves the best results among all the classifiers. 
It indicates benefits of the synthetic labeled images are beyond the simple SDANN. 
These observations demonstrate the effectiveness of ADDES in labeled images generation with inexact supervision.

\textbf{Visualization}: Samples of synthetic data on MJSynth-1 are presented in Figure \ref{fig:vis_image}. The target label set is $\{e\}$. The first row are samples generated with $\{e\}$ set to $1$ and the second row are samples with  $\{e\}$ set to $0$. We could observe that ADDES generate realistic data to facilitate the training of classifiers for target label prediction.

\subsection{Impacts of the Size of $\mathcal{D}_l$}

\begin{figure}[t]
\centering
\vspace{-1mm}
\begin{subfigure}{0.49\columnwidth}
    \centering
    \includegraphics[width=\linewidth]{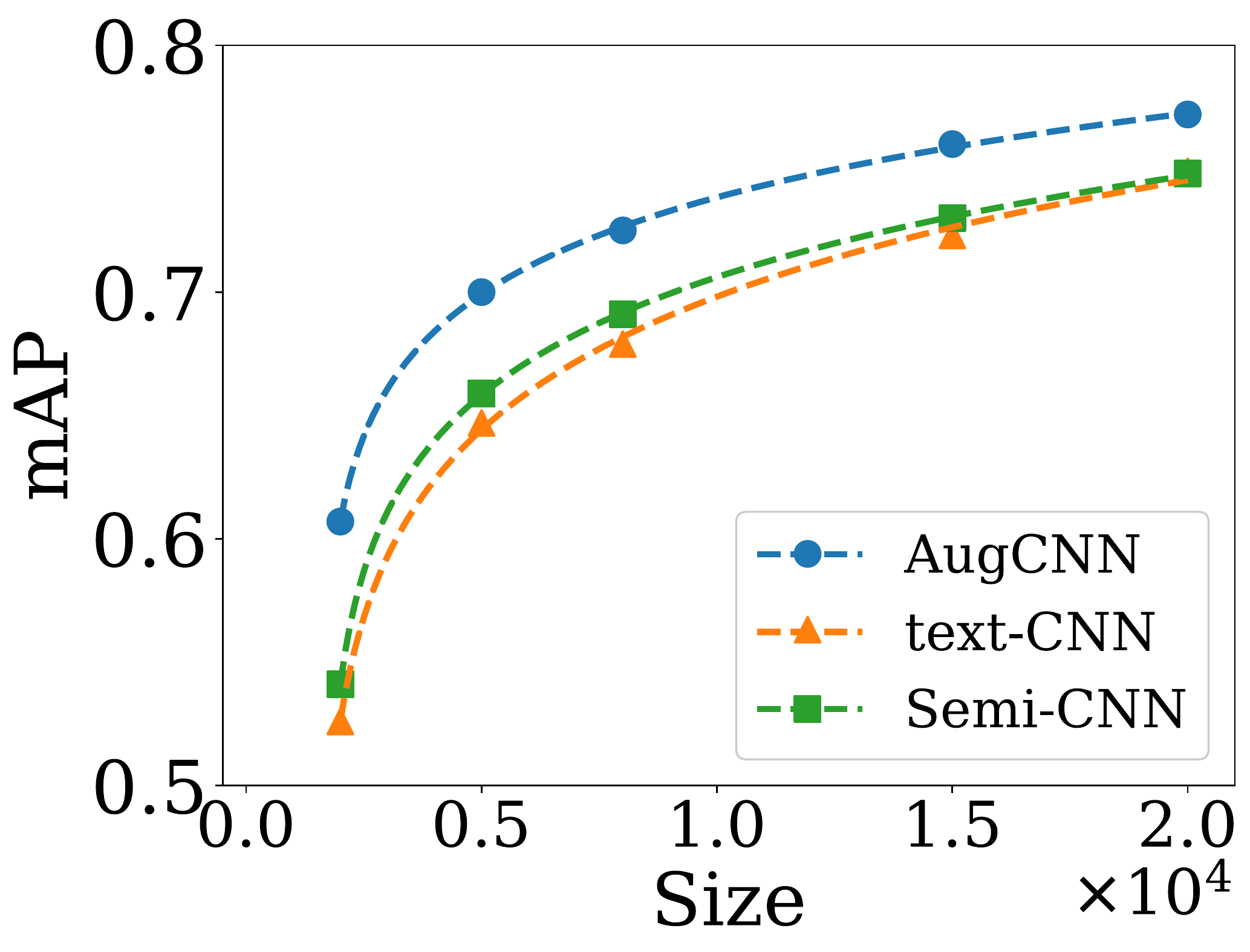} 
    \vspace{-7mm}
    \caption{mAP on StackOverflow-1}
    \label{fig:label_text_ap}
\end{subfigure}
\begin{subfigure}{0.49\columnwidth}
    \centering
    \includegraphics[width=\linewidth]{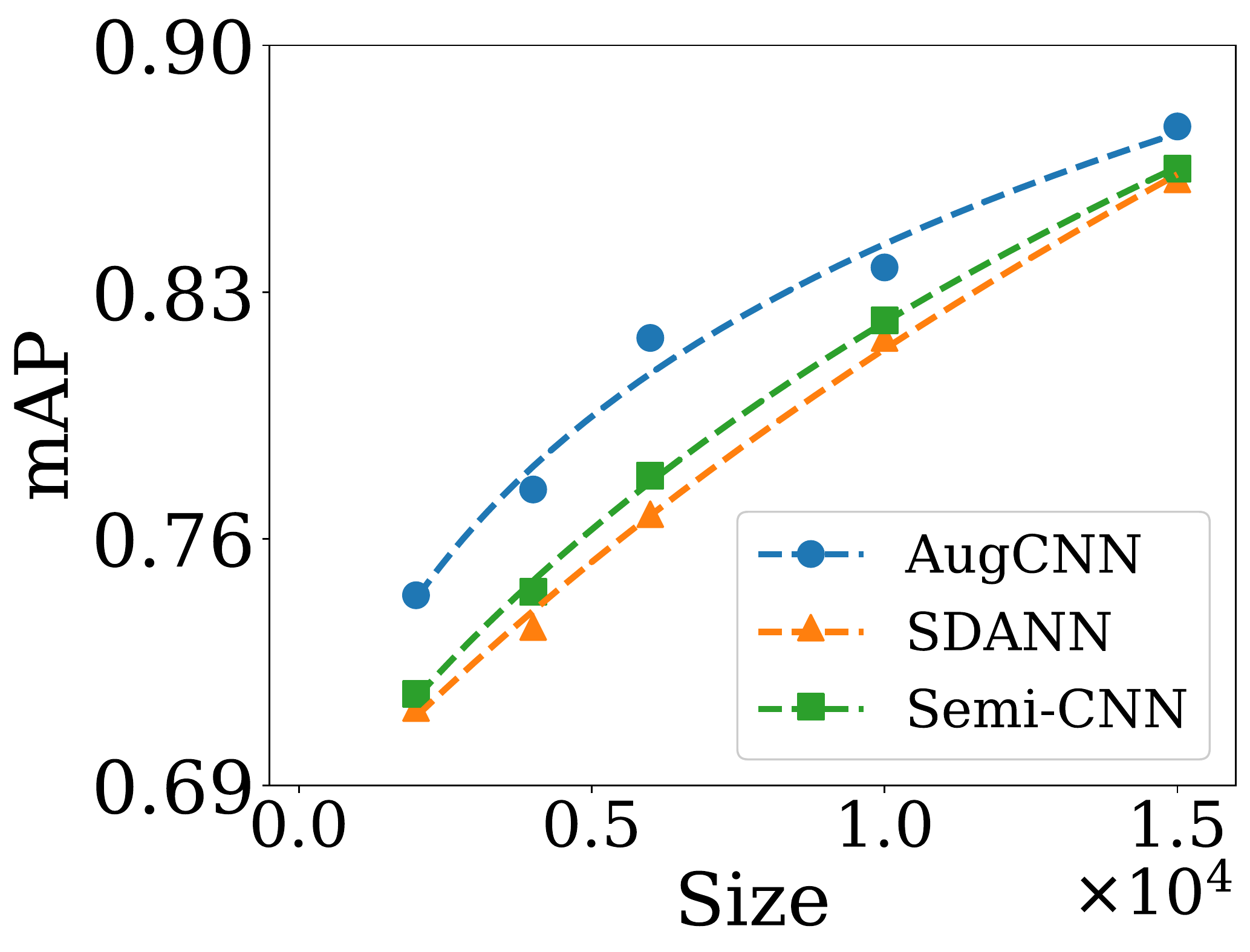} 
    \vspace{-7mm}
    \caption{mAP on MJSynth-1}
    \label{fig:label_text_auc}
\end{subfigure}
\vskip -1em
\caption{Impacts of the size of $\mathcal{D}_l$ to our proposed model.}
\vspace{-3mm}
\label{fig:label_curve}
\end{figure}

Prior studies have shown that some semi-supervised learning methods and data augmentation methods are sensitive to the size of labeled data \cite{oliver2018realistic,xie2019unsupervised}.
To demonstrate our proposed model could synthesize useful labeled data to facilitate the targets label prediction regardless of the size of $\mathcal{D}_l$, we conduct experiments on StackOverflow-1 and MJSynth-1 to answer \textbf{RQ2}. We select the sizes of $\mathcal{D}_l$ ranging from 2k to 20k. The results are shown in Figure \ref{fig:label_curve}. To make fair comparisons, we compare text-CNN/SDANN, Semi-CNN, and AugCNN, which have the same network structure. From Figure \ref{fig:label_curve}, we could observe that in both text and image datasets, Semi-CNN makes negligible improvements compared with text-CNN. On the 
contrary, with the synthetic labeled data included in the training set, AugCNN consistently outperforms the other baselines with a clear margin regardless of the size of $\mathcal{D}_l$. It shows that our proposed method could 
benefit the scenarios varying in data types and sizes.

\begin{figure}[h]
\centering
\begin{subfigure}{0.45\columnwidth}
    \centering
    \includegraphics[width=\linewidth]{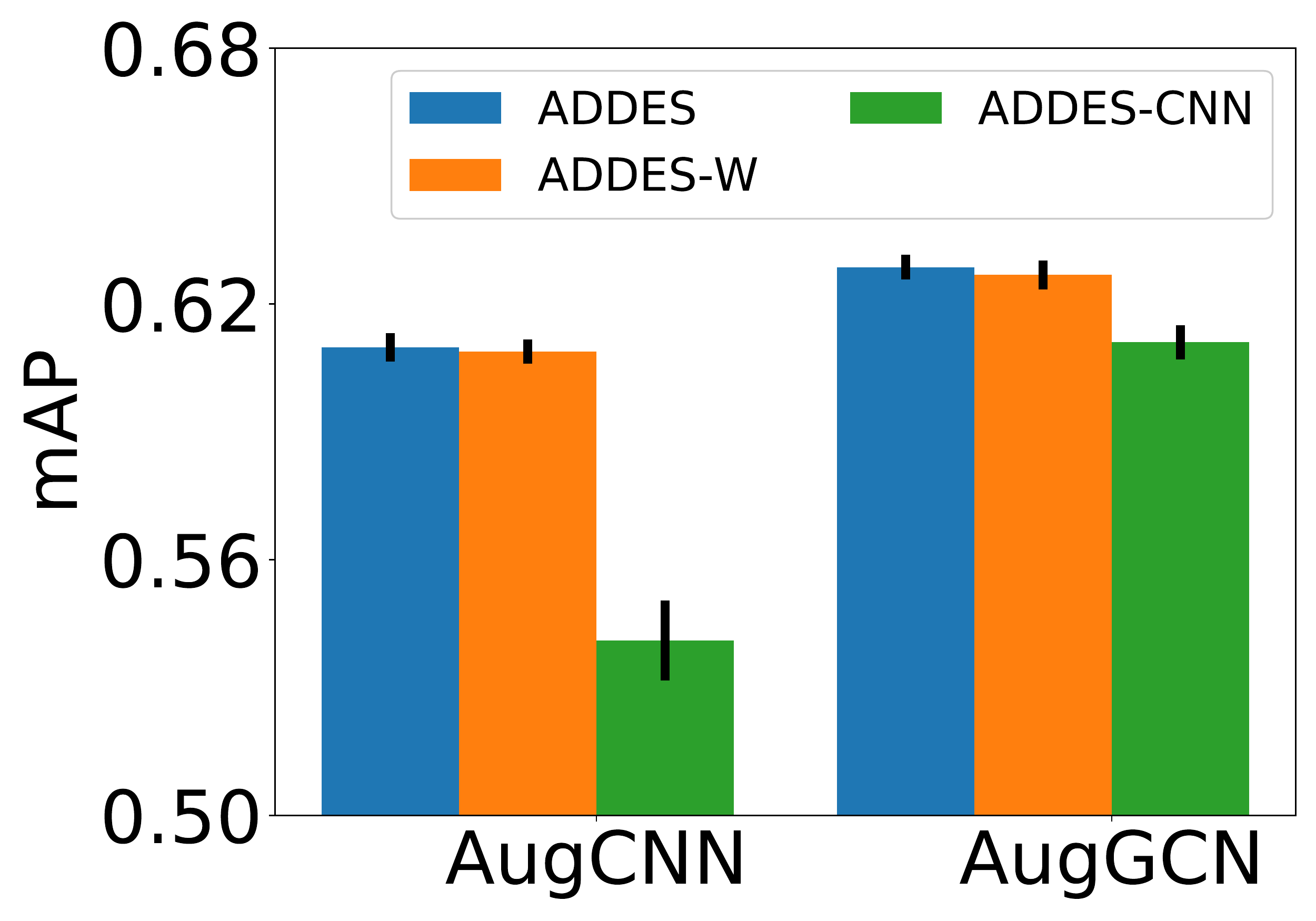} 
    \vspace{-5mm}
    \caption{mAP}
    \label{fig:graph_ap}
\end{subfigure}
\begin{subfigure}{0.45\columnwidth}
    \centering
    \includegraphics[width=\linewidth]{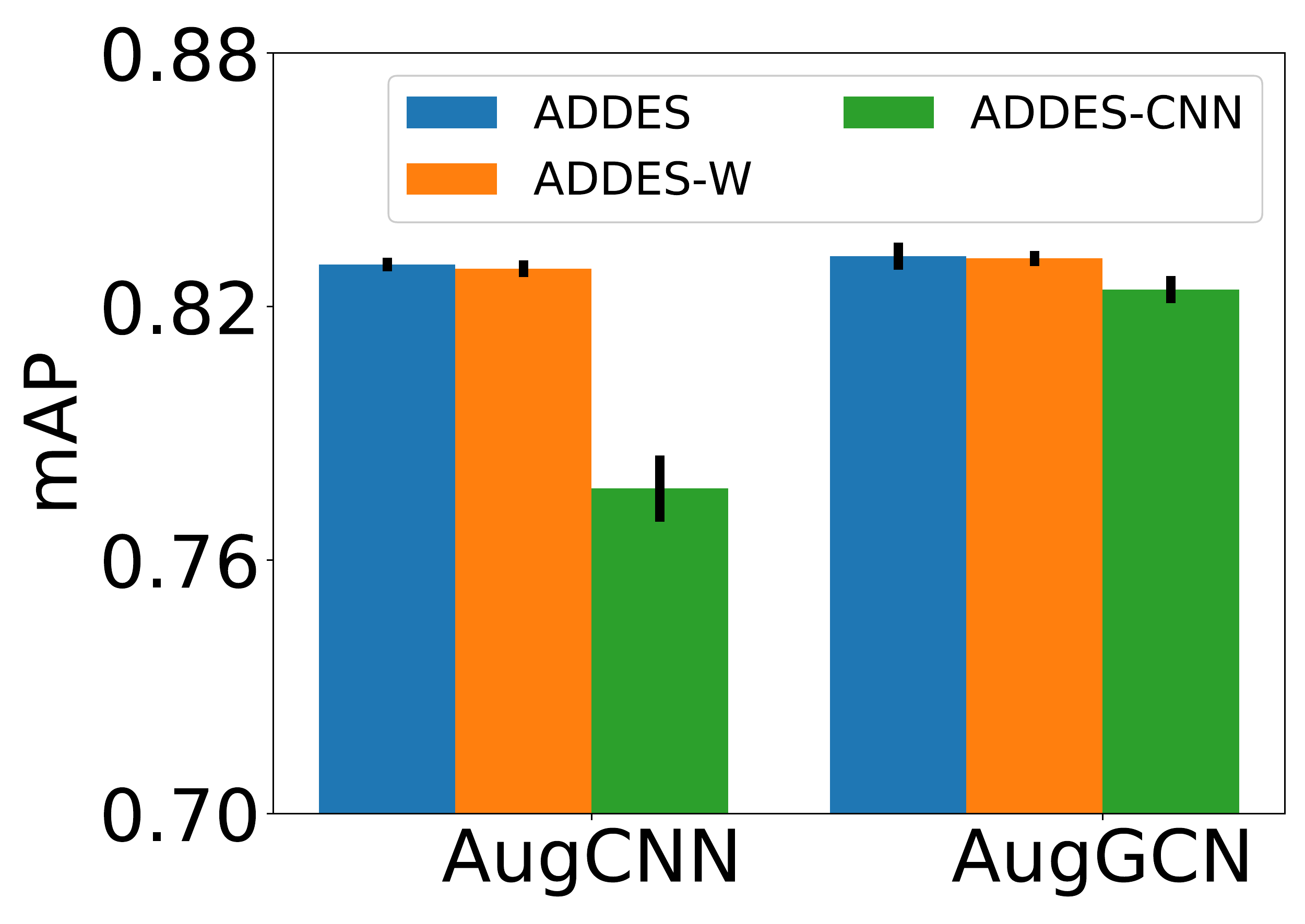} 
    \vspace{-5mm}
    \caption{AUC}
    \label{fig:graph_auc}
\end{subfigure}
\vskip -1.5em
\caption{Comparisons with ADDES and its variants.}
\vspace{-3mm}
\label{fig:graph_compare}
\end{figure}

\subsection{Ablation Study }

To answer \textbf{RQ3}, we conduct ablation studies on StackOverflow-1 to investigate the importance of the GCN-based classifier and its sensitivity to the graph construction methods.
Specifically, we compare our model with the following variants of ADDES:
\begin{itemize}[leftmargin=*]
    \item \textbf{ADDES-CNN}: It replaces the GCN-based classifier in ADDES with a multi-label classifier to obtain $q_C(\mathbf{y}_t,\mathbf{y}_s|\mathbf{x})$. It treats the prediction of multiple labels as isolation tasks.
    \item \textbf{ADDES-W}: ADDES-W builds weighted graph thorough the ground truth of training data in whole class set through Eq.(\ref{eq:correlation}) for the GCN-based classifier. The weights of edges between classes indicate their co-occurrence rates.   
\end{itemize}
The performance of the AugCNN and AugGCN which utilize 30k synthetic data from ADDES and its variants is presented in Figure \ref{fig:graph_compare}. As we can see, if we eliminate the GCN module, the gain brought by the synthetic data will significantly decrease ($p<0.005$, t-test). However, we could find that the synthetic data of ADDES and ADDES-W shows no significant difference for data augmentation. 
From these observations, we could conclude that (1) the information aggregation from the inexact supervision to target labels contributes to better generative model for inexact supervision; (2) The graph utilizing prior knowledge to obtain binary weights between target classes and inexpensive supervision classes shows no difference with the graph completely built by the labels co-occurrence probability in modeling the data with inexact supervision.



\begin{figure}[h]
\centering
\vskip -1em
\begin{subfigure}{0.48\columnwidth}
    \centering
    \includegraphics[width=\linewidth]{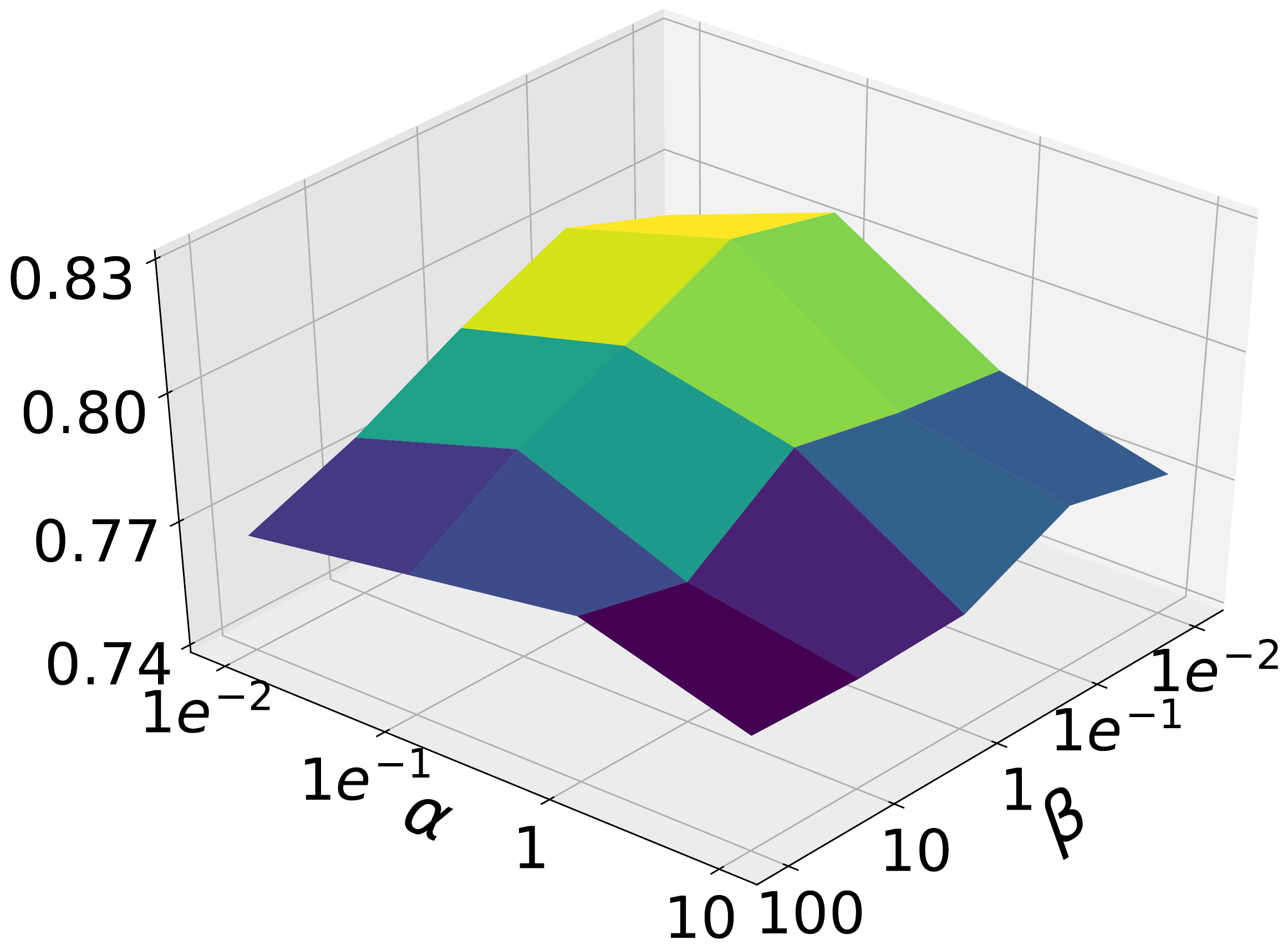} 
    \vspace{-5mm}
    \caption{mAP of AugCNN}
    \label{fig:para_map}
\end{subfigure}
\begin{subfigure}{0.48\columnwidth}
    \centering
    \includegraphics[width=\linewidth]{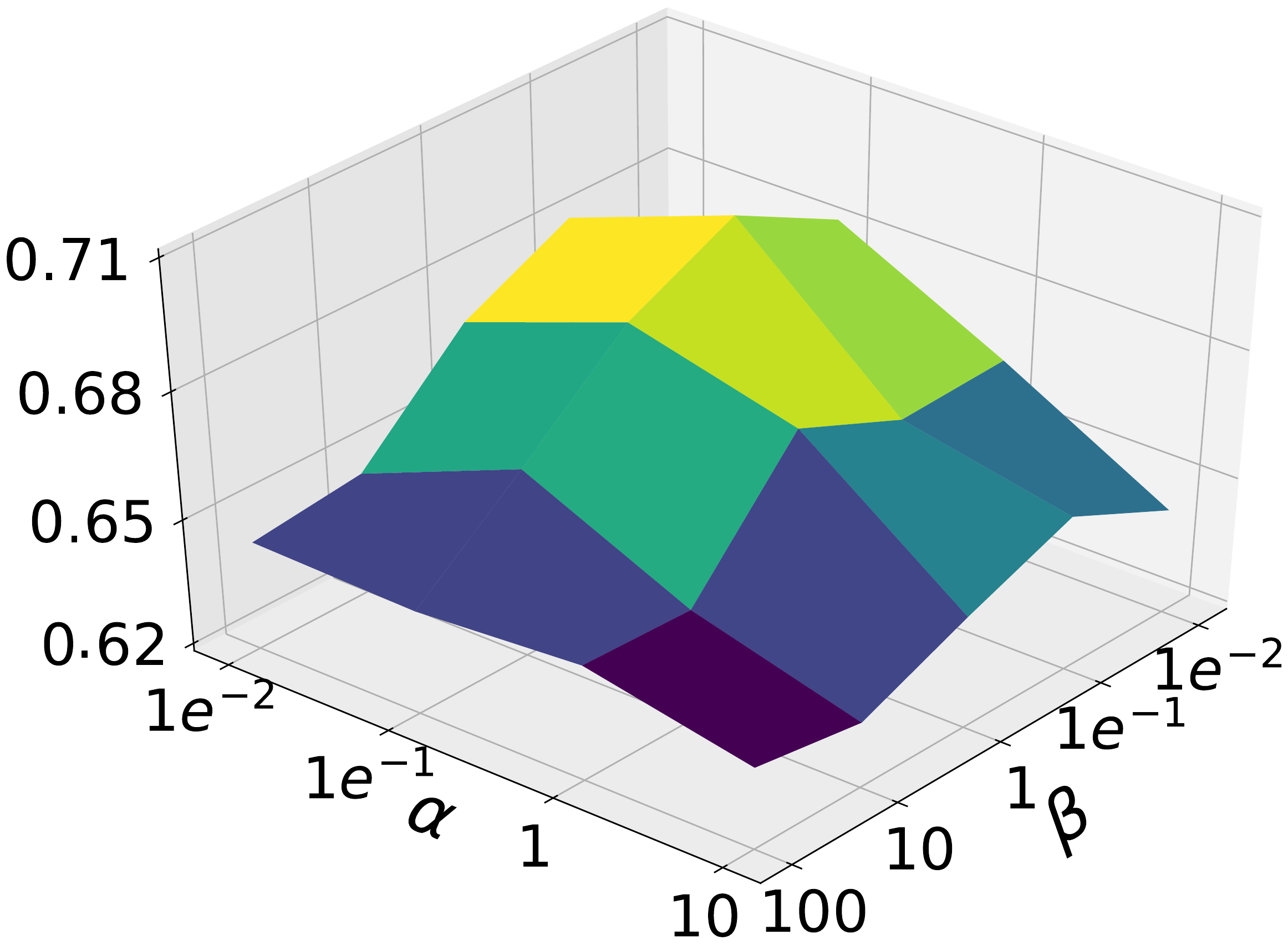} 
    \vspace{-5mm}
    \caption{AUC of AugCNN}
    \label{fig:para_auc}
\end{subfigure}
\vskip -1em
\caption{Parameter sensitivity analysis.}
\vskip -1em
\label{fig:para}
\end{figure}

\subsection{Parameter Sensitivity}
The proposed framework includes two important hyperparameters, i.e., $\alpha$ controlling the contribution of the constraint for disentangled representation learning, $\beta$ controlling the contribution of the inexact supervision labels to model the classifier in ADDES. We investigate the impacts of these two parameters on target label prediction on MJSynth-1 with the number of synthetic data set as 30k. We vary $\alpha$ as $\{0.01,0.1,1,10\}$ and $\beta$ as $\{0.01,0.1,1,10,100\}$. Then, We obtain AugCNN models with the synthetic data from the generative models. The results are presented in Fig. \ref{fig:para}. With the increase of $\alpha$, the performance first increase then decrease. The same trend also exhibits in $\beta$. And when both $\alpha$ and $\beta$ ranges from 0.01 to 1, the generative model shows consistently good performance.

\section{conclusion and future work} \label{sec:conclusion}
In this paper, we investigate a novel problem of labeled data generation with inexact supervision. It is a potential direction to cope with the deficiency of labeled data for deep learning. To deal with this problem, we propose a novel generative framework ADDES to generate data labeled in both target and inexact supervision class set. Extensive experimental results on image and text datasets demonstrated the effectiveness of the ADDES in  synthesizing high-quality labeled data for the target label prediction. Further experiments are conducted to understand the contributions of each component of ADDES and its parameter sensitivity.  There are several interesting directions which need further investigation. First, in this paper, we assume the 
inexact supervision labels are clean. However, the labels could be noisy as they are crawled from social media. Thus, one direction is to investigate labeled data generation with inexact and inaccurate supervision. Second, there are many different ways in constructing the graph. We would like to study  automatic methods to construct the graph linking the related labels. 

\section{Acknowledgements}
This material is based upon work supported by, or in part by, the National Science Foundation (NSF) under grant  \#IIS-1909702, \#IIS1955851. The findings and conclusions in this paper do not necessarily reflect the view of the funding agency.

\bibliographystyle{ACM-Reference-Format}
\bibliography{ref}


\begin{thebibliography}{46}


\ifx \showCODEN    \undefined \def \showCODEN     #1{\unskip}     \fi
\ifx \showDOI      \undefined \def \showDOI       #1{#1}\fi
\ifx \showISBNx    \undefined \def \showISBNx     #1{\unskip}     \fi
\ifx \showISBNxiii \undefined \def \showISBNxiii  #1{\unskip}     \fi
\ifx \showISSN     \undefined \def \showISSN      #1{\unskip}     \fi
\ifx \showLCCN     \undefined \def \showLCCN      #1{\unskip}     \fi
\ifx \shownote     \undefined \def \shownote      #1{#1}          \fi
\ifx \showarticletitle \undefined \def \showarticletitle #1{#1}   \fi
\ifx \showURL      \undefined \def \showURL       {\relax}        \fi
\providecommand\bibfield[2]{#2}
\providecommand\bibinfo[2]{#2}
\providecommand\natexlab[1]{#1}
\providecommand\showeprint[2][]{arXiv:#2}

\bibitem[\protect\citeauthoryear{Antoniou, Storkey, and Edwards}{Antoniou
  et~al\mbox{.}}{2017}]%
        {antoniou2017data}
\bibfield{author}{\bibinfo{person}{Antreas Antoniou}, \bibinfo{person}{Amos
  Storkey}, {and} \bibinfo{person}{Harrison Edwards}.}
  \bibinfo{year}{2017}\natexlab{}.
\newblock \showarticletitle{Data augmentation generative adversarial networks}.
\newblock \bibinfo{journal}{\emph{arXiv preprint arXiv:1711.04340}}
  (\bibinfo{year}{2017}).
\newblock


\bibitem[\protect\citeauthoryear{Boutell, Luo, Shen, and Brown}{Boutell
  et~al\mbox{.}}{2004}]%
        {boutell2004learning}
\bibfield{author}{\bibinfo{person}{Matthew~R Boutell}, \bibinfo{person}{Jiebo
  Luo}, \bibinfo{person}{Xipeng Shen}, {and} \bibinfo{person}{Christopher~M
  Brown}.} \bibinfo{year}{2004}\natexlab{}.
\newblock \showarticletitle{Learning multi-label scene classification}.
\newblock \bibinfo{journal}{\emph{Pattern recognition}} \bibinfo{volume}{37},
  \bibinfo{number}{9} (\bibinfo{year}{2004}), \bibinfo{pages}{1757--1771}.
\newblock


\bibitem[\protect\citeauthoryear{Bowman, Vilnis, Vinyals, Dai, Jozefowicz, and
  Bengio}{Bowman et~al\mbox{.}}{2015}]%
        {bowman2015generating}
\bibfield{author}{\bibinfo{person}{Samuel~R Bowman}, \bibinfo{person}{Luke
  Vilnis}, \bibinfo{person}{Oriol Vinyals}, \bibinfo{person}{Andrew~M Dai},
  \bibinfo{person}{Rafal Jozefowicz}, {and} \bibinfo{person}{Samy Bengio}.}
  \bibinfo{year}{2015}\natexlab{}.
\newblock \showarticletitle{Generating sentences from a continuous space}.
\newblock \bibinfo{journal}{\emph{arXiv preprint arXiv:1511.06349}}
  (\bibinfo{year}{2015}).
\newblock


\bibitem[\protect\citeauthoryear{Chen, Wei, Wang, and Guo}{Chen
  et~al\mbox{.}}{2019}]%
        {chen_multi-label_2019}
\bibfield{author}{\bibinfo{person}{Zhao-Min Chen}, \bibinfo{person}{Xiu-Shen
  Wei}, \bibinfo{person}{Peng Wang}, {and} \bibinfo{person}{Yanwen Guo}.}
  \bibinfo{year}{2019}\natexlab{}.
\newblock \showarticletitle{Multi-Label Image Recognition with Graph
  Convolutional Networks}. In \bibinfo{booktitle}{\emph{CVPR}}.
  \bibinfo{pages}{5177--5186}.
\newblock


\bibitem[\protect\citeauthoryear{Cho, Van~Merri{\"e}nboer, Gulcehre, Bahdanau,
  Bougares, Schwenk, and Bengio}{Cho et~al\mbox{.}}{2014}]%
        {cho2014learning}
\bibfield{author}{\bibinfo{person}{Kyunghyun Cho}, \bibinfo{person}{Bart
  Van~Merri{\"e}nboer}, \bibinfo{person}{Caglar Gulcehre},
  \bibinfo{person}{Dzmitry Bahdanau}, \bibinfo{person}{Fethi Bougares},
  \bibinfo{person}{Holger Schwenk}, {and} \bibinfo{person}{Yoshua Bengio}.}
  \bibinfo{year}{2014}\natexlab{}.
\newblock \showarticletitle{Learning phrase representations using RNN
  encoder-decoder for statistical machine translation}.
\newblock \bibinfo{journal}{\emph{arXiv preprint arXiv:1406.1078}}
  (\bibinfo{year}{2014}).
\newblock


\bibitem[\protect\citeauthoryear{Frome, Corrado, Shlens, Bengio, Dean, Ranzato,
  and Mikolov}{Frome et~al\mbox{.}}{2013}]%
        {frome2013devise}
\bibfield{author}{\bibinfo{person}{Andrea Frome}, \bibinfo{person}{Greg~S
  Corrado}, \bibinfo{person}{Jon Shlens}, \bibinfo{person}{Samy Bengio},
  \bibinfo{person}{Jeff Dean}, \bibinfo{person}{Marc'Aurelio Ranzato}, {and}
  \bibinfo{person}{Tomas Mikolov}.} \bibinfo{year}{2013}\natexlab{}.
\newblock \showarticletitle{Devise: A deep visual-semantic embedding model}. In
  \bibinfo{booktitle}{\emph{NeurIPS}}. \bibinfo{pages}{2121--2129}.
\newblock


\bibitem[\protect\citeauthoryear{Gehring, Auli, Grangier, Yarats, and
  Dauphin}{Gehring et~al\mbox{.}}{2017}]%
        {gehring2017convolutional}
\bibfield{author}{\bibinfo{person}{Jonas Gehring}, \bibinfo{person}{Michael
  Auli}, \bibinfo{person}{David Grangier}, \bibinfo{person}{Denis Yarats},
  {and} \bibinfo{person}{Yann~N Dauphin}.} \bibinfo{year}{2017}\natexlab{}.
\newblock \showarticletitle{Convolutional sequence to sequence learning}. In
  \bibinfo{booktitle}{\emph{ICML}}. \bibinfo{pages}{1243--1252}.
\newblock


\bibitem[\protect\citeauthoryear{Goodfellow, Pouget-Abadie, Mirza, Xu,
  Warde-Farley, Ozair, Courville, and Bengio}{Goodfellow et~al\mbox{.}}{2014}]%
        {goodfellow2014generative}
\bibfield{author}{\bibinfo{person}{Ian Goodfellow}, \bibinfo{person}{Jean
  Pouget-Abadie}, \bibinfo{person}{Mehdi Mirza}, \bibinfo{person}{Bing Xu},
  \bibinfo{person}{David Warde-Farley}, \bibinfo{person}{Sherjil Ozair},
  \bibinfo{person}{Aaron Courville}, {and} \bibinfo{person}{Yoshua Bengio}.}
  \bibinfo{year}{2014}\natexlab{}.
\newblock \showarticletitle{Generative adversarial nets}. In
  \bibinfo{booktitle}{\emph{NeurIPS}}. \bibinfo{pages}{2672--2680}.
\newblock


\bibitem[\protect\citeauthoryear{Grandvalet and Bengio}{Grandvalet and
  Bengio}{2005}]%
        {grandvalet2005semi}
\bibfield{author}{\bibinfo{person}{Yves Grandvalet} {and}
  \bibinfo{person}{Yoshua Bengio}.} \bibinfo{year}{2005}\natexlab{}.
\newblock \showarticletitle{Semi-supervised learning by entropy minimization}.
  In \bibinfo{booktitle}{\emph{NeurIPS}}. \bibinfo{pages}{529--536}.
\newblock


\bibitem[\protect\citeauthoryear{Han, Luo, and Wang}{Han et~al\mbox{.}}{2019}]%
        {han2019deep}
\bibfield{author}{\bibinfo{person}{Jiangfan Han}, \bibinfo{person}{Ping Luo},
  {and} \bibinfo{person}{Xiaogang Wang}.} \bibinfo{year}{2019}\natexlab{}.
\newblock \showarticletitle{Deep self-learning from noisy labels}. In
  \bibinfo{booktitle}{\emph{CVPR}}. \bibinfo{pages}{5138--5147}.
\newblock


\bibitem[\protect\citeauthoryear{He, Zhang, Ren, and Sun}{He
  et~al\mbox{.}}{2016}]%
        {he2016deep}
\bibfield{author}{\bibinfo{person}{Kaiming He}, \bibinfo{person}{Xiangyu
  Zhang}, \bibinfo{person}{Shaoqing Ren}, {and} \bibinfo{person}{Jian Sun}.}
  \bibinfo{year}{2016}\natexlab{}.
\newblock \showarticletitle{Deep residual learning for image recognition}. In
  \bibinfo{booktitle}{\emph{CVPR}}. \bibinfo{pages}{770--778}.
\newblock


\bibitem[\protect\citeauthoryear{Hochreiter and Schmidhuber}{Hochreiter and
  Schmidhuber}{1997}]%
        {hochreiter1997long}
\bibfield{author}{\bibinfo{person}{Sepp Hochreiter} {and}
  \bibinfo{person}{J{\"u}rgen Schmidhuber}.} \bibinfo{year}{1997}\natexlab{}.
\newblock \showarticletitle{Long short-term memory}.
\newblock \bibinfo{journal}{\emph{Neural computation}} \bibinfo{volume}{9},
  \bibinfo{number}{8} (\bibinfo{year}{1997}), \bibinfo{pages}{1735--1780}.
\newblock


\bibitem[\protect\citeauthoryear{Hu, Yang, Liang, Salakhutdinov, and Xing}{Hu
  et~al\mbox{.}}{2017}]%
        {hu2017toward}
\bibfield{author}{\bibinfo{person}{Zhiting Hu}, \bibinfo{person}{Zichao Yang},
  \bibinfo{person}{Xiaodan Liang}, \bibinfo{person}{Ruslan Salakhutdinov},
  {and} \bibinfo{person}{Eric~P Xing}.} \bibinfo{year}{2017}\natexlab{}.
\newblock \showarticletitle{Toward controlled generation of text}. In
  \bibinfo{booktitle}{\emph{ICML}}. JMLR. org, \bibinfo{pages}{1587--1596}.
\newblock


\bibitem[\protect\citeauthoryear{Jaderberg, Simonyan, Vedaldi, and
  Zisserman}{Jaderberg et~al\mbox{.}}{2014}]%
        {Jaderberg14c}
\bibfield{author}{\bibinfo{person}{Max Jaderberg}, \bibinfo{person}{Karen
  Simonyan}, \bibinfo{person}{Andrea Vedaldi}, {and} \bibinfo{person}{Andrew
  Zisserman}.} \bibinfo{year}{2014}\natexlab{}.
\newblock \showarticletitle{Synthetic Data and Artificial Neural Networks for
  Natural Scene Text Recognition}. In \bibinfo{booktitle}{\emph{Workshop on
  Deep Learning, NIPS}}.
\newblock


\bibitem[\protect\citeauthoryear{Kampffmeyer, Chen, Liang, Wang, Zhang, and
  Xing}{Kampffmeyer et~al\mbox{.}}{2019}]%
        {Kampffmeyer_2019_CVPR}
\bibfield{author}{\bibinfo{person}{Michael Kampffmeyer}, \bibinfo{person}{Yinbo
  Chen}, \bibinfo{person}{Xiaodan Liang}, \bibinfo{person}{Hao Wang},
  \bibinfo{person}{Yujia Zhang}, {and} \bibinfo{person}{Eric~P. Xing}.}
  \bibinfo{year}{2019}\natexlab{}.
\newblock \showarticletitle{Rethinking Knowledge Graph Propagation for
  Zero-Shot Learning}. In \bibinfo{booktitle}{\emph{CVPR}}.
\newblock


\bibitem[\protect\citeauthoryear{Kim}{Kim}{2014}]%
        {kim2014convolutional}
\bibfield{author}{\bibinfo{person}{Yoon Kim}.} \bibinfo{year}{2014}\natexlab{}.
\newblock \showarticletitle{Convolutional neural networks for sentence
  classification}.
\newblock \bibinfo{journal}{\emph{arXiv preprint arXiv:1408.5882}}
  (\bibinfo{year}{2014}).
\newblock


\bibitem[\protect\citeauthoryear{Kingma, Mohamed, Rezende, and Welling}{Kingma
  et~al\mbox{.}}{2014}]%
        {kingma2014semi}
\bibfield{author}{\bibinfo{person}{Durk~P Kingma}, \bibinfo{person}{Shakir
  Mohamed}, \bibinfo{person}{Danilo~Jimenez Rezende}, {and}
  \bibinfo{person}{Max Welling}.} \bibinfo{year}{2014}\natexlab{}.
\newblock \showarticletitle{Semi-supervised learning with deep generative
  models}. In \bibinfo{booktitle}{\emph{NeurIPS}}. \bibinfo{pages}{3581--3589}.
\newblock


\bibitem[\protect\citeauthoryear{Kingma and Welling}{Kingma and
  Welling}{2013}]%
        {kingma2013autoencoding}
\bibfield{author}{\bibinfo{person}{Diederik~P Kingma} {and}
  \bibinfo{person}{Max Welling}.} \bibinfo{year}{2013}\natexlab{}.
\newblock \bibinfo{title}{Auto-Encoding Variational Bayes}.
\newblock
\newblock
\showeprint[arxiv]{stat.ML/1312.6114}


\bibitem[\protect\citeauthoryear{Krizhevsky, Sutskever, and Hinton}{Krizhevsky
  et~al\mbox{.}}{2012}]%
        {krizhevsky2012imagenet}
\bibfield{author}{\bibinfo{person}{Alex Krizhevsky}, \bibinfo{person}{Ilya
  Sutskever}, {and} \bibinfo{person}{Geoffrey~E Hinton}.}
  \bibinfo{year}{2012}\natexlab{}.
\newblock \showarticletitle{Imagenet classification with deep convolutional
  neural networks}. In \bibinfo{booktitle}{\emph{NeurIPS}}.
  \bibinfo{pages}{1097--1105}.
\newblock


\bibitem[\protect\citeauthoryear{Kumar~Verma, Arora, Mishra, and
  Rai}{Kumar~Verma et~al\mbox{.}}{2018}]%
        {Verma_2018_CVPR}
\bibfield{author}{\bibinfo{person}{Vinay Kumar~Verma}, \bibinfo{person}{Gundeep
  Arora}, \bibinfo{person}{Ashish Mishra}, {and} \bibinfo{person}{Piyush Rai}.}
  \bibinfo{year}{2018}\natexlab{}.
\newblock \showarticletitle{Generalized Zero-Shot Learning via Synthesized
  Examples}. In \bibinfo{booktitle}{\emph{CVPR}}.
\newblock


\bibitem[\protect\citeauthoryear{Lee, Fang, Yeh, and Wang}{Lee
  et~al\mbox{.}}{2018}]%
        {lee_multi-label_2018}
\bibfield{author}{\bibinfo{person}{Chung-Wei Lee}, \bibinfo{person}{Wei Fang},
  \bibinfo{person}{Chih-Kuan Yeh}, {and} \bibinfo{person}{Yu-Chiang~Frank
  Wang}.} \bibinfo{year}{2018}\natexlab{}.
\newblock \showarticletitle{Multi-label zero-shot learning with structured
  knowledge graphs}. In \bibinfo{booktitle}{\emph{CVPR}}.
  \bibinfo{pages}{1576--1585}.
\newblock


\bibitem[\protect\citeauthoryear{Luong, Pham, and Manning}{Luong
  et~al\mbox{.}}{2015}]%
        {luong2015effective}
\bibfield{author}{\bibinfo{person}{Minh-Thang Luong}, \bibinfo{person}{Hieu
  Pham}, {and} \bibinfo{person}{Christopher~D Manning}.}
  \bibinfo{year}{2015}\natexlab{}.
\newblock \showarticletitle{Effective approaches to attention-based neural
  machine translation}.
\newblock \bibinfo{journal}{\emph{arXiv preprint arXiv:1508.04025}}
  (\bibinfo{year}{2015}).
\newblock


\bibitem[\protect\citeauthoryear{Mirza and Osindero}{Mirza and
  Osindero}{2014}]%
        {mirza2014conditional}
\bibfield{author}{\bibinfo{person}{Mehdi Mirza} {and} \bibinfo{person}{Simon
  Osindero}.} \bibinfo{year}{2014}\natexlab{}.
\newblock \showarticletitle{Conditional generative adversarial nets}.
\newblock \bibinfo{journal}{\emph{arXiv preprint arXiv:1411.1784}}
  (\bibinfo{year}{2014}).
\newblock


\bibitem[\protect\citeauthoryear{Mishra, Krishna~Reddy, Mittal, and
  Murthy}{Mishra et~al\mbox{.}}{2018}]%
        {Mishra_2018_CVPR_Workshops}
\bibfield{author}{\bibinfo{person}{Ashish Mishra}, \bibinfo{person}{Shiva
  Krishna~Reddy}, \bibinfo{person}{Anurag Mittal}, {and}
  \bibinfo{person}{Hema~A. Murthy}.} \bibinfo{year}{2018}\natexlab{}.
\newblock \showarticletitle{A Generative Model for Zero Shot Learning Using
  Conditional Variational Autoencoders}. In \bibinfo{booktitle}{\emph{CVPR
  Workshops}}.
\newblock


\bibitem[\protect\citeauthoryear{Odena, Olah, and Shlens}{Odena
  et~al\mbox{.}}{2017}]%
        {odena2017conditional}
\bibfield{author}{\bibinfo{person}{Augustus Odena},
  \bibinfo{person}{Christopher Olah}, {and} \bibinfo{person}{Jonathon Shlens}.}
  \bibinfo{year}{2017}\natexlab{}.
\newblock \showarticletitle{Conditional image synthesis with auxiliary
  classifier gans}. In \bibinfo{booktitle}{\emph{ICML}}. JMLR. org,
  \bibinfo{pages}{2642--2651}.
\newblock


\bibitem[\protect\citeauthoryear{Oliver, Odena, Raffel, Cubuk, and
  Goodfellow}{Oliver et~al\mbox{.}}{2018}]%
        {oliver2018realistic}
\bibfield{author}{\bibinfo{person}{Avital Oliver}, \bibinfo{person}{Augustus
  Odena}, \bibinfo{person}{Colin~A Raffel}, \bibinfo{person}{Ekin~Dogus Cubuk},
  {and} \bibinfo{person}{Ian Goodfellow}.} \bibinfo{year}{2018}\natexlab{}.
\newblock \showarticletitle{Realistic evaluation of deep semi-supervised
  learning algorithms}. In \bibinfo{booktitle}{\emph{NeurIPS}}.
  \bibinfo{pages}{3235--3246}.
\newblock


\bibitem[\protect\citeauthoryear{Qin, Xu, and Wang}{Qin et~al\mbox{.}}{2018}]%
        {qin2018dsgan}
\bibfield{author}{\bibinfo{person}{Pengda Qin}, \bibinfo{person}{Weiran Xu},
  {and} \bibinfo{person}{William~Yang Wang}.} \bibinfo{year}{2018}\natexlab{}.
\newblock \showarticletitle{Dsgan: Generative adversarial training for distant
  supervision relation extraction}.
\newblock \bibinfo{journal}{\emph{arXiv preprint arXiv:1805.09929}}
  (\bibinfo{year}{2018}).
\newblock


\bibitem[\protect\citeauthoryear{Ratner, Bach, Ehrenberg, Fries, Wu, and
  R{\'e}}{Ratner et~al\mbox{.}}{2017}]%
        {ratner2017snorkel}
\bibfield{author}{\bibinfo{person}{Alexander Ratner},
  \bibinfo{person}{Stephen~H Bach}, \bibinfo{person}{Henry Ehrenberg},
  \bibinfo{person}{Jason Fries}, \bibinfo{person}{Sen Wu}, {and}
  \bibinfo{person}{Christopher R{\'e}}.} \bibinfo{year}{2017}\natexlab{}.
\newblock \showarticletitle{Snorkel: Rapid training data creation with weak
  supervision}. In \bibinfo{booktitle}{\emph{VLDB}}, Vol.~\bibinfo{volume}{11}.
  \bibinfo{pages}{269}.
\newblock


\bibitem[\protect\citeauthoryear{Ren, He, Girshick, and Sun}{Ren
  et~al\mbox{.}}{2015}]%
        {ren2015faster}
\bibfield{author}{\bibinfo{person}{Shaoqing Ren}, \bibinfo{person}{Kaiming He},
  \bibinfo{person}{Ross Girshick}, {and} \bibinfo{person}{Jian Sun}.}
  \bibinfo{year}{2015}\natexlab{}.
\newblock \showarticletitle{Faster r-cnn: Towards real-time object detection
  with region proposal networks}. In \bibinfo{booktitle}{\emph{NeurIPS}}.
  \bibinfo{pages}{91--99}.
\newblock


\bibitem[\protect\citeauthoryear{Romera-Paredes and Torr}{Romera-Paredes and
  Torr}{2015}]%
        {romera2015embarrassingly}
\bibfield{author}{\bibinfo{person}{Bernardino Romera-Paredes} {and}
  \bibinfo{person}{Philip Torr}.} \bibinfo{year}{2015}\natexlab{}.
\newblock \showarticletitle{An embarrassingly simple approach to zero-shot
  learning}. In \bibinfo{booktitle}{\emph{ICML}}. \bibinfo{pages}{2152--2161}.
\newblock


\bibitem[\protect\citeauthoryear{Shin, Tenenholtz, Rogers, Schwarz, Senjem,
  Gunter, Andriole, and Michalski}{Shin et~al\mbox{.}}{[n.d.]}]%
        {shin2018medical}
\bibfield{author}{\bibinfo{person}{Hoo-Chang Shin}, \bibinfo{person}{Neil~A
  Tenenholtz}, \bibinfo{person}{Jameson~K Rogers},
  \bibinfo{person}{Christopher~G Schwarz}, \bibinfo{person}{Matthew~L Senjem},
  \bibinfo{person}{Jeffrey~L Gunter}, \bibinfo{person}{Katherine~P Andriole},
  {and} \bibinfo{person}{Mark Michalski}.} \bibinfo{year}{[n.d.]}\natexlab{}.
\newblock \showarticletitle{Medical image synthesis for data augmentation and
  anonymization using generative adversarial networks}. In
  \bibinfo{booktitle}{\emph{International workshop on simulation and synthesis
  in medical imaging}}. \bibinfo{pages}{1--11}.
\newblock


\bibitem[\protect\citeauthoryear{Shu, Wang, Le, Lee, and Liu}{Shu
  et~al\mbox{.}}{2018}]%
        {shu2018deep}
\bibfield{author}{\bibinfo{person}{Kai Shu}, \bibinfo{person}{Suhang Wang},
  \bibinfo{person}{Thai Le}, \bibinfo{person}{Dongwon Lee}, {and}
  \bibinfo{person}{Huan Liu}.} \bibinfo{year}{2018}\natexlab{}.
\newblock \showarticletitle{Deep headline generation for clickbait detection}.
  In \bibinfo{booktitle}{\emph{ICDM}}. IEEE, \bibinfo{pages}{467--476}.
\newblock


\bibitem[\protect\citeauthoryear{Simonyan and Zisserman}{Simonyan and
  Zisserman}{2014}]%
        {simonyan2014very}
\bibfield{author}{\bibinfo{person}{Karen Simonyan} {and}
  \bibinfo{person}{Andrew Zisserman}.} \bibinfo{year}{2014}\natexlab{}.
\newblock \showarticletitle{Very deep convolutional networks for large-scale
  image recognition}.
\newblock \bibinfo{journal}{\emph{arXiv preprint arXiv:1409.1556}}
  (\bibinfo{year}{2014}).
\newblock


\bibitem[\protect\citeauthoryear{Sohn, Lee, and Yan}{Sohn
  et~al\mbox{.}}{2015}]%
        {sohn2015learning}
\bibfield{author}{\bibinfo{person}{Kihyuk Sohn}, \bibinfo{person}{Honglak Lee},
  {and} \bibinfo{person}{Xinchen Yan}.} \bibinfo{year}{2015}\natexlab{}.
\newblock \showarticletitle{Learning structured output representation using
  deep conditional generative models}. In \bibinfo{booktitle}{\emph{NeurIPS}}.
  \bibinfo{pages}{3483--3491}.
\newblock


\bibitem[\protect\citeauthoryear{Vaswani, Shazeer, Parmar, Uszkoreit, Jones,
  Gomez, Kaiser, and Polosukhin}{Vaswani et~al\mbox{.}}{2017}]%
        {vaswani2017attention}
\bibfield{author}{\bibinfo{person}{Ashish Vaswani}, \bibinfo{person}{Noam
  Shazeer}, \bibinfo{person}{Niki Parmar}, \bibinfo{person}{Jakob Uszkoreit},
  \bibinfo{person}{Llion Jones}, \bibinfo{person}{Aidan~N Gomez},
  \bibinfo{person}{{\L}ukasz Kaiser}, {and} \bibinfo{person}{Illia
  Polosukhin}.} \bibinfo{year}{2017}\natexlab{}.
\newblock \showarticletitle{Attention is all you need}. In
  \bibinfo{booktitle}{\emph{NeurIPS}}. \bibinfo{pages}{5998--6008}.
\newblock


\bibitem[\protect\citeauthoryear{Veit, Alldrin, Chechik, Krasin, Gupta, and
  Belongie}{Veit et~al\mbox{.}}{2017}]%
        {veit2017learning}
\bibfield{author}{\bibinfo{person}{Andreas Veit}, \bibinfo{person}{Neil
  Alldrin}, \bibinfo{person}{Gal Chechik}, \bibinfo{person}{Ivan Krasin},
  \bibinfo{person}{Abhinav Gupta}, {and} \bibinfo{person}{Serge Belongie}.}
  \bibinfo{year}{2017}\natexlab{}.
\newblock \showarticletitle{Learning from noisy large-scale datasets with
  minimal supervision}. In \bibinfo{booktitle}{\emph{CVPR}}.
  \bibinfo{pages}{839--847}.
\newblock


\bibitem[\protect\citeauthoryear{Wang, Yang, Mao, Huang, Huang, and Xu}{Wang
  et~al\mbox{.}}{2016}]%
        {wang2016cnn}
\bibfield{author}{\bibinfo{person}{Jiang Wang}, \bibinfo{person}{Yi Yang},
  \bibinfo{person}{Junhua Mao}, \bibinfo{person}{Zhiheng Huang},
  \bibinfo{person}{Chang Huang}, {and} \bibinfo{person}{Wei Xu}.}
  \bibinfo{year}{2016}\natexlab{}.
\newblock \showarticletitle{Cnn-rnn: A unified framework for multi-label image
  classification}. In \bibinfo{booktitle}{\emph{CVPR}}.
  \bibinfo{pages}{2285--2294}.
\newblock


\bibitem[\protect\citeauthoryear{Wang, Derr, Ma, Wang, Liu, Liu, and Tang}{Wang
  et~al\mbox{.}}{2020a}]%
        {wang2020learning}
\bibfield{author}{\bibinfo{person}{Wentao Wang}, \bibinfo{person}{Tyler Derr},
  \bibinfo{person}{Yao Ma}, \bibinfo{person}{Suhang Wang}, \bibinfo{person}{Hui
  Liu}, \bibinfo{person}{Zitao Liu}, {and} \bibinfo{person}{Jiliang Tang}.}
  \bibinfo{year}{2020}\natexlab{a}.
\newblock \showarticletitle{Learning from Incomplete Labeled Data via
  Adversarial Data Generation}. In \bibinfo{booktitle}{\emph{ICDM}}. IEEE,
  \bibinfo{pages}{1316--1321}.
\newblock


\bibitem[\protect\citeauthoryear{Wang, Wang, Fan, Liu, and Tang}{Wang
  et~al\mbox{.}}{2020b}]%
        {wang2020global}
\bibfield{author}{\bibinfo{person}{Wentao Wang}, \bibinfo{person}{Suhang Wang},
  \bibinfo{person}{Wenqi Fan}, \bibinfo{person}{Zitao Liu}, {and}
  \bibinfo{person}{Jiliang Tang}.} \bibinfo{year}{2020}\natexlab{b}.
\newblock \showarticletitle{Global-and-Local Aware Data Generation for the
  Class Imbalance Problem}. In \bibinfo{booktitle}{\emph{SDM}}. SIAM,
  \bibinfo{pages}{307--315}.
\newblock


\bibitem[\protect\citeauthoryear{Wang, Ye, and Gupta}{Wang
  et~al\mbox{.}}{2018b}]%
        {Wang_2018_CVPR}
\bibfield{author}{\bibinfo{person}{Xiaolong Wang}, \bibinfo{person}{Yufei Ye},
  {and} \bibinfo{person}{Abhinav Gupta}.} \bibinfo{year}{2018}\natexlab{b}.
\newblock \showarticletitle{Zero-Shot Recognition via Semantic Embeddings and
  Knowledge Graphs}. In \bibinfo{booktitle}{\emph{CVPR}}.
\newblock


\bibitem[\protect\citeauthoryear{Wang, Wang, Qi, Tang, and Li}{Wang
  et~al\mbox{.}}{2018a}]%
        {wang2018weakly}
\bibfield{author}{\bibinfo{person}{Yilin Wang}, \bibinfo{person}{Suhang Wang},
  \bibinfo{person}{Guojun Qi}, \bibinfo{person}{Jiliang Tang}, {and}
  \bibinfo{person}{Baoxin Li}.} \bibinfo{year}{2018}\natexlab{a}.
\newblock \showarticletitle{Weakly supervised facial attribute manipulation via
  deep adversarial network}. In \bibinfo{booktitle}{\emph{WACV}}. IEEE,
  \bibinfo{pages}{112--121}.
\newblock


\bibitem[\protect\citeauthoryear{Xian, Sharma, Schiele, and Akata}{Xian
  et~al\mbox{.}}{2019}]%
        {xian2019f}
\bibfield{author}{\bibinfo{person}{Yongqin Xian}, \bibinfo{person}{Saurabh
  Sharma}, \bibinfo{person}{Bernt Schiele}, {and} \bibinfo{person}{Zeynep
  Akata}.} \bibinfo{year}{2019}\natexlab{}.
\newblock \showarticletitle{f-VAEGAN-D2: A feature generating framework for
  any-shot learning}. In \bibinfo{booktitle}{\emph{CVPR}}.
  \bibinfo{pages}{10275--10284}.
\newblock


\bibitem[\protect\citeauthoryear{Xiao, Xia, Yang, Huang, and Wang}{Xiao
  et~al\mbox{.}}{2015}]%
        {xiao2015learning}
\bibfield{author}{\bibinfo{person}{Tong Xiao}, \bibinfo{person}{Tian Xia},
  \bibinfo{person}{Yi Yang}, \bibinfo{person}{Chang Huang}, {and}
  \bibinfo{person}{Xiaogang Wang}.} \bibinfo{year}{2015}\natexlab{}.
\newblock \showarticletitle{Learning from massive noisy labeled data for image
  classification}. In \bibinfo{booktitle}{\emph{CVPR}}.
  \bibinfo{pages}{2691--2699}.
\newblock


\bibitem[\protect\citeauthoryear{Xie, Dai, Hovy, Luong, and Le}{Xie
  et~al\mbox{.}}{2019}]%
        {xie2019unsupervised}
\bibfield{author}{\bibinfo{person}{Qizhe Xie}, \bibinfo{person}{Zihang Dai},
  \bibinfo{person}{Eduard Hovy}, \bibinfo{person}{Minh-Thang Luong}, {and}
  \bibinfo{person}{Quoc~V Le}.} \bibinfo{year}{2019}\natexlab{}.
\newblock \showarticletitle{Unsupervised data augmentation for consistency
  training}.
\newblock  (\bibinfo{year}{2019}).
\newblock


\bibitem[\protect\citeauthoryear{Zhang, Xiang, and Gong}{Zhang
  et~al\mbox{.}}{2017}]%
        {zhang2017learning}
\bibfield{author}{\bibinfo{person}{Li Zhang}, \bibinfo{person}{Tao Xiang},
  {and} \bibinfo{person}{Shaogang Gong}.} \bibinfo{year}{2017}\natexlab{}.
\newblock \showarticletitle{Learning a deep embedding model for zero-shot
  learning}. In \bibinfo{booktitle}{\emph{CVPR}}. \bibinfo{pages}{2021--2030}.
\newblock


\bibitem[\protect\citeauthoryear{Zhou}{Zhou}{2018}]%
        {zhou2018brief}
\bibfield{author}{\bibinfo{person}{Zhi-Hua Zhou}.}
  \bibinfo{year}{2018}\natexlab{}.
\newblock \showarticletitle{A brief introduction to weakly supervised
  learning}.
\newblock \bibinfo{journal}{\emph{National science review}}
  \bibinfo{volume}{5}, \bibinfo{number}{1} (\bibinfo{year}{2018}),
  \bibinfo{pages}{44--53}.
\newblock


\end{thebibliography}

\end{document}